\documentclass[11pt,a4paper]{article}
\usepackage{amsmath,amssymb,calc,ifthen}
\usepackage{float}
\usepackage[table,usenames,dvipsnames]{xcolor} 
\usepackage{tikz}
\usepackage{hyperref}
\usepackage{url}
\hypersetup{
colorlinks,
citecolor=black,
filecolor=black,
linkcolor=black,
urlcolor=black
}
\usetikzlibrary{plotmarks,shapes}
\usepackage{amsmath,graphicx}
\usepackage{mathtools}
\usepackage{epstopdf}
\usepackage{caption}
\usepackage{subcaption}
\usepackage{graphicx}
\usepackage{color}

\newcommand\ci{\perp\!\!\!\perp} 

\usepackage{listings}
\usepackage[margin=0.5in]{geometry}
\usepackage{pdfpages}
\usepackage{enumitem} 

\usepackage{titlesec} 

\usepackage{pdfpages} 

\DeclareMathOperator*{\argmin}{arg\,min}
\DeclareMathOperator*{\argmax}{arg\,max}

\title{Supplementary Material}
\date{}

\begin{document}
\belowdisplayskip=12pt plus 3pt minus 9pt
\belowdisplayshortskip=7pt plus 3pt minus 4pt



# DIVE: A spatiotemporal progression model of brain pathology in neurodegenerative disorders

Răzvan V. Marinescu[1], Arman Eshaghi[2], Marco Lorenzi[1,3], Alexandra L. Young[1], Neil P. Oxtoby[1], Sara Garbarino[1,3], Sebastian J. Crutch[4], Daniel C. Alexander[1], for the Alzheimer's Disease Neuroimaging Initiative

1. Centre for Medical Image Computing, University College London, Gower Street, London, United Kingdom
2. Queen Square Multiple Sclerosis Centre, UCL Institute of Neurology, Faculty of Brain Sciences, University College London, United Kingdom
3. Epione Research-Project, Inria, France and Université Côte d'Azur, France
4. Dementia Research Centre, University College London Institute of Neurology, London, United Kingdom

**Corresponding author:**
Răzvan Valentin Marinescu
Email: razvan.marinescu.14@ucl.ac.uk
Address: Engineering Front Building, Torrington Place, London, WC1E 7JE

## Abstract

Current models of progression in neurodegenerative diseases use neuroimaging measures that are averaged across pre-defined regions of interest (ROIs). Such models are unable to recover fine details of atrophy patterns; they tend to impose an assumption of strong spatial correlation within each ROI and no correlation among ROIs. Such assumptions may be violated by the influence of underlying brain network connectivity on pathology propagation – a strong hypothesis e.g. in Alzheimer's Disease. Here we present DIVE: **D**ata-driven **I**nference of **V**ertexwise **E**volution. DIVE is an image-based disease progression model with single-vertex resolution, designed to reconstruct long-term patterns of brain pathology from short-term longitudinal data sets. DIVE clusters vertex-wise (i.e. point-wise) biomarker measurements on the cortical surface that have similar temporal dynamics across a patient population, and concurrently estimates an average trajectory of vertex measurements in each cluster. DIVE uniquely outputs a parcellation of the cortex into areas with common progression patterns, leading to a new signature for individual diseases. DIVE further estimates the disease stage and progression speed for every visit of every subject, potentially enhancing stratification for clinical trials or management. On simulated data, DIVE can recover ground truth clusters and their underlying trajectory, provided the average trajectories are sufficiently different between clusters. We demonstrate DIVE on data from two cohorts: the Alzheimer's Disease Neuroimaging Initiative (ADNI) and the Dementia Research Centre (DRC), UK. The DRC cohort contains patients with Posterior Cortical Atrophy (PCA) as well as typical Alzheimer's disease (tAD). DIVE finds similar spatial patterns of atrophy

for tAD subjects in the two independent datasets (ADNI and DRC), and further reveals distinct patterns of pathology in different diseases (tAD vs PCA) and for distinct types of biomarker data – cortical thickness from Magnetic Resonance Imaging (MRI) vs amyloid load from Positron Emission Tomography (PET). We demonstrate that DIVE stages have potential clinical relevance, despite being based only on imaging data, by showing that the stages correlate with cognitive test scores. Finally, DIVE can be used to estimate a fine-grained spatial distribution of pathology in the brain using any kind of voxelwise or vertexwise measures including Jacobian compression maps, fractional anisotropy (FA) maps from diffusion tensor imaging (DTI) or other PET measures.

DIVE availability: DIVE source code, written in Python3, is available at https://github.com/mrazvan22/dive and can be easily applied on any registered voxelwise images or images processed with the Freesurfer software. ADNI data can be downloaded from the Laboratory of NeuroImaging at the University of Southern California.

# 1 Introduction

Many biomarkers exist that can be used to track the severity of neurodegenerative diseases such as Alzheimer's disease (AD). Clinical function can be measured using cognitive assessments performed by an expert clinician and brain atrophy can be measured using Magnetic Resonance Imaging (MRI). Other measures include molecular markers such as aggregation of misfolded amyloid-beta or tau measured using Positron Emission Tomography (PET), and measures of white-matter degradation such as fractional anisotropy (FA) from Diffusion Tensor Imaging (DTI). The evolution of these biomarkers across the disease time-course creates a unique signature of the disease that can be used to stage patients, which is helpful for stratification in clinical trials.

A hypothetical model of disease progression has been proposed by (Jack Jr et al., 2010), describing the order of abnormality of key biomarkers along the progression of AD. The model suggests that amyloid-beta and tau biomarkers become abnormal long before symptoms appear, followed by brain atrophy measures and cognitive decline. Motivated by this hypothetical model, several data-driven disease progression models have been proposed in recent years, which aggregate information from multiple biomarkers into a single time frame representing disease progression. One such model is the Event-Based Model (EBM) (Fonteijn et al., 2012; Young et al., 2014), which models the progression of disease as a sequence of discrete events, representing underlying biomarkers switching from a normal to abnormal state. Other types of disease progression models (Donohue et al., 2014; Jedynak et al., 2012; Li et al., 2017; Lorenzi et al., 2017; Schiratti et al., 2015) have been developed, that build continuous trajectories by "stitching" together short-term follow-up data from

individual subjects. In contrast to the discrete disease stages that are estimated by the EBM, these models also compute a continuous measure of disease stage for every individual by estimating individual time shifts and progression speeds.

Current image-based disease progression models estimate the evolution of the disease using a small set of biomarkers corresponding to pre-defined regions-of-interest (ROI). This ROI parcellation is usually coarse, doesn't allow one to find spatially dispersed patterns of atrophy. While spatiotemporal longitudinal models have already been demonstrated (Derado et al., 2010; Hyun et al., 2016; Lorenzi et al., 2015), these models regress against pre-defined sets of covariates such as age or clinical markers. This is problematic because, in the case of age, this assumes all subjects have the same age of disease onset. Similarly, clinical markers are noisy, biased, suffer from floor/ceiling effects and are not sensitive in pre-symptomatic phases. Recently, some spatiotemporal models that estimate subject-specific time-shifts have been developed (Bilgel et al., 2016; Koval et al., 2017). However, these models generally cannot recover dispersed and disconnected pathological patterns, because they assume correlation of voxels based on spatial distance, either through a distance function or distance from control points. However, spatially dispersed pathological patterns have been observed in AD and related dementias and are hypothesized to appear due to interactions of pathology and brain networks (Seeley et al., 2009). Therefore, a spatiotemporal disease progression model that allows recovery of dispersed atrophy patterns present in AD, is not currently available.

In this work, we present DIVE: **D**ata-driven **I**nference of **V**ertexwise **E**volution. DIVE is a novel disease progression model with single vertex resolution that makes only weak assumptions on spatial correlation. In contrast to approaches which model temporal trajectories for a small set of biomarker measures based on *a priori* defined ROIs, DIVE models temporal trajectories for each vertex on the cortical surface. DIVE combines unsupervised learning and disease progression modelling to identify clusters of vertices on the cortical surface that show a similar trajectory of brain pathology over a particular patient cohort. This formulation enables us to estimate a fine-grained spatial distribution of pathology and also provides a novel parcellation of the brain based on temporal change.

We first test DIVE on synthetic data and show that the model can recover known biomarker trajectories and disease progression scores. We then demonstrate the model on data from two cohorts: the Alzheimer's Disease Neuroimaging Initiative (ADNI) and the Dementia Research Centre (DRC), UK. We use the model to reveal spatiotemporal patterns of pathology to a much finer resolution than previous models and demonstrate the ability to assign subjects to stages that predict progression. Finally, we validate DIVE in terms of how robust are the estimated pathology patterns and how well the disease progression scores correlate with cognitive tests.

# 2 Method

In this section we describe the mathematical formulation of DIVE (section 2.1), then we show how to fit the model using Expectation Maximisation (section 2.2) and we describe further implementation details of the algorithm (section 2.3). Afterwards, we outline the synthetic data-generation process (section 2.4) for testing the model in the presence of ground truth, as well as the pipeline for pre-processing the ADNI and DRC datasets (section 2.5).

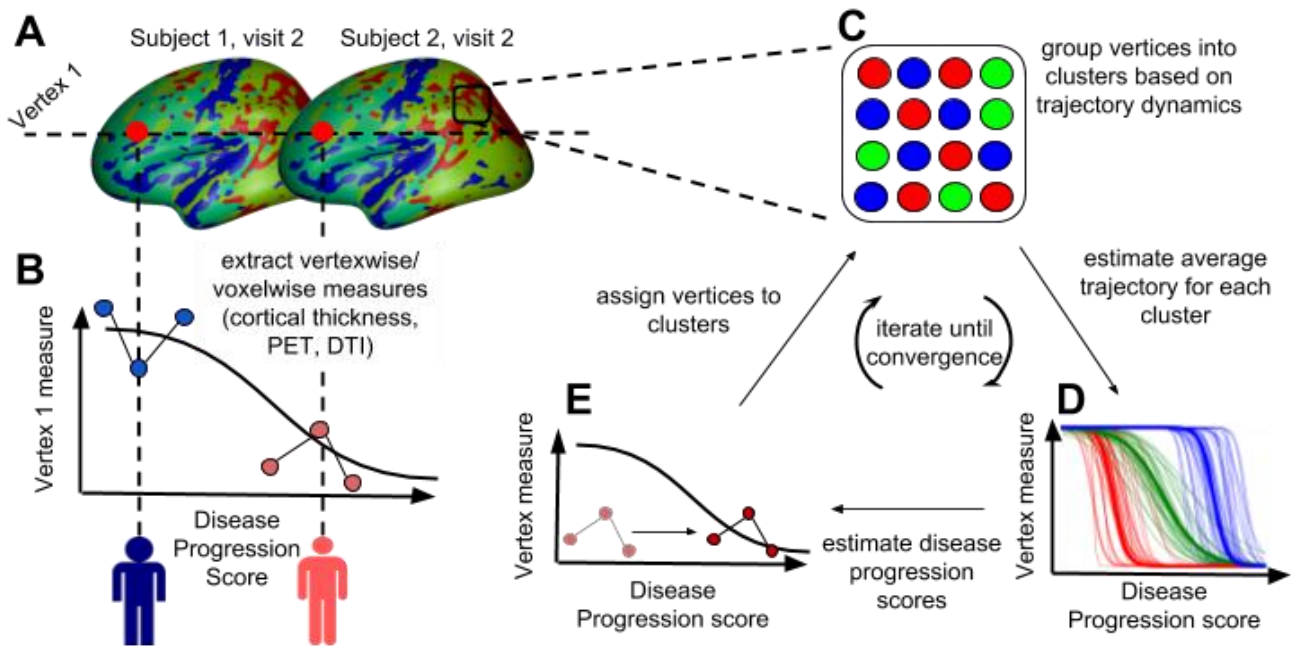

Figure 1. Diagram of the proposed DIVE model. DIVE assumes that biomarkers of pathology (e.g. cortical thinning) can be measured at many vertices (i.e. locations) on the cortical surface (**A**), where each vertex has a distinct trajectory of change during disease progression (**B**). The model thus assigns to every cortical vertex one of a small set of temporal trajectories describing the change in some image-based measurement (e.g. cortical thickness, amyloid PET, DTI fractional anisotropy measures) from beginning to end of the disease progression. The estimation process simultaneously estimates the set of clusters, the trajectory defining each cluster, and the position of each subject along the trajectories, which are defined on a common time line. The process iterates assignment of each vertex to clusters (red, green and blue in this diagram) (**C**), estimation of the trajectory in each cluster (**D**) and estimation of the disease progression score (location along trajectory) for each subject (**E**), all within an expectation-maximisation framework, until convergence.

## 2.1 The DIVE model

Figure 1 illustrates the DIVE aims and implementation. DIVE input measures are vertexwise or voxelwise biomarker measures in the brain (Fig 1A), such as cortical

thickness or amyloid load. A vertex is a location on the cortical surface at which a biomarker of pathology is quantifiable (e.g. cortical thickness). For each vertex on the cortical surface (or voxel in the 3D brain volume), we estimate a unique trajectory along the disease progression timeline (Fig 1B), while also estimating subject/visit-specific disease progression scores (i.e. disease stages). We do that by grouping vertices with similar biomarker trajectories into clusters (Fig 1C), and we estimate a representative trajectory for every cluster (Fig 1D). Each trajectory is a function of subject-/visit-specific disease progression scores (DPS) (Fig 1E). The DPS depends linearly on the time since baseline visit, but with subject-specific slope and intercept.

### 2.1.1 Modelling Subject-specific Time Shifts

The disease progression score $s_{ij}$ for subject $i$ at visit $j$ is defined as a linear transformation of time since baseline measurement $t_{ij}$ (in years):

$$s_{ij} = \alpha_i t_{ij} + \beta_i \tag{1}$$

where $\alpha_i$ and $\beta_i$ represent the speed of progression and time shift (i.e. disease onset) of subject $i$.

### 2.1.2 Modelling Biomarker Trajectory for a Single Vertex

DIVE assumes that the biomarker measure at each vertex on the cortical surface follows a sigmoidal trajectory $f( . ; \theta)$ over the disease progression score $s$ and with parameters $\theta$. We choose a parametric sigmoid function because it is a parsimonious parametric model that offers better fit compared to linear models, is monotonic, and can account for floor and ceiling effects (Caroli and Frisoni, 2010; Sabuncu et al., 2011). We also assume that vertices are grouped into $K$ clusters and we model a unique trajectory for each cluster $k \in \{1, ... , K\}$, which will be referred to as cluster trajectories. The sigmoidal function $f(s; \theta_k)$ for cluster $k$ is defined as:

$$f(s; \theta_k) = \frac{a_k}{1 + exp(-b_k(s - c_k))} + d_k \tag{2}$$

where $s$ is the disease progression score from Eq. 1 and $\theta_k = [a_k, b_k, c_k, d_k]$ are parameters controlling the shape of the trajectory – $d_k$ and $d_k+a_k$ represent the lower and upper limits of the sigmoidal function, $c_k$ represents the inflection point and $a_k b_k/4$ represents the slope at the inflection point. For a given subject $i$ at visit $j$, the value $V_l^{ij}$ of its biomarker measurement at vertex $l$ is a random variable that has an associated discrete latent variable $Z_l \in [1, ... , K]$ denoting the cluster it was generated from. The value of $V_l^{ij}$ given that it was generated from cluster $Z_l$ can be modelled as:

$$p(V_l^{ij}|\alpha_i, \beta_i, \theta_{Z_l}, \sigma_{Z_l}, Z_l) = N(V_l^{ij}|f(\alpha_i t_{ij} + \beta_i|\theta_{Z_l}), \sigma_{Z_l}) \tag{3}$$

where $N(V_l^{ij} | f(\alpha_i t_{ij} + \beta_i | \theta_{Z_l}), \sigma_{Z_l})$ represents the probability density function (pdf) of the normal distribution that models the measurement noise along the sigmoidal trajectory of cluster $Z_l$, having variance $\sigma_{Z_l}$. Next, we assume the measurements from different subjects are independent, while the measurements from the same subject $i$ at different visits $j$ are linked using the disease progression score from Eq. 1. Moreover, we also assume a uniform prior on cluster membership $Z_l$. This gives the following model:

$$p(V_l, Z_l | \alpha, \beta, \theta, \sigma) = \prod_{(i,j) \in I} N(V_l^{ij} | f(\alpha_i t_{ij} + \beta_i | \theta_{Z_l}), \sigma_{Z_l}) \quad (4)$$

where $I = \{(i, j)\}$ represents the set of all the subjects $i$ and their corresponding visits $j$. Furthermore, $V_l = [V_l^{ij} | (i,j) \in I]$ is the 1D array of all the values for vertex $l$ across every visit of every subject. Vectors $\alpha = [\alpha_1, ... , \alpha_S]$ and $\beta = [\beta_1, ... , \beta_S]$, where $S$ is the number of subjects, denote the stacked parameters for the subject shifts. If a subject $i$ has multiple visits, these visits share the same parameters $\alpha_i$ and $\beta_i$. Vectors $\theta = [\theta_1, ... , \theta_K]$ and $\sigma = [\sigma_1, ... , \sigma_K]$, with $K$ being the number of clusters, represent the stacked parameters for the sigmoidal trajectories and measurement noise specific to each cluster.

### 2.1.3 Modelling Biomarker Trajectories for all Vertices

So far we have a model for only one vertex on the brain surface. We therefore extend the formulation to all the vertices by assuming all these vertex measurements are spatially independent, giving the complete data likelihood:

$$p(V, Z | \alpha, \beta, \theta, \sigma) = \prod_l^L \prod_{(i,j) \in I} N(V_l^{ij} | f(\alpha_i t_{ij} + \beta_i | \theta_{Z_l}), \sigma_{Z_l}) \quad (5)$$

where $V = [V_1, ... , V_l]$, $Z = [Z_1, ... , Z_l]$, $L$ being the total number of vertices on the cortical surface. The formulation assumes spatial independence between measurements in different vertices, but in section 2.1.4 the model is extended to capture spatial correlations. We get the final model log likelihood for incomplete data by marginalising over the latent variables $Z$:

$$p(V | \alpha, \beta, \theta, \sigma) = \prod_{l=1}^{L} \sum_{k=1}^{K} p(Z_l = k) \prod_{(i,j) \in I} N(V_l^{ij} | f(\alpha_i t_{ij} + \beta_i | \theta_k), \sigma_k) \quad (6)$$

Throughout the article, we will use the shorthand $z_{lk} = p(Z_l = k)$.

### 2.1.4 Modelling Spatial Correlation

The version of the model presented so far does not make any assumptions on spatial correlation. However, the regional organisation of the cortex suggests we would expect spatial correlation of the vertex measurements. More precisely, measures of cortical thickness or other modalities are often similar in neighbouring vertices on the cortical surface and likely belong to the same cluster. DIVE can be easily extended to include mild spatial constraints on the correlation of vertex measurements via a Markov Random Field (MRF), which encourages neighbouring vertices to have the same corresponding cluster. We hypothesise that incorporating such constraints should reduce the effects of noise and produce a more stable clustering. However, this does not model correlation between the actual vertex values, but only between the latent variables $Z_l$, i.e. the cluster membership of each vertex. The MRF thus has the advantage of not requiring the use of huge covariance matrices, which are otherwise needed if we want to model correlation of vertex values directly. With the MRF, the full-data likelihood function of the model now becomes:

$$p(V, Z|\alpha, \beta, \theta, \sigma, \lambda) = \prod_l^L \left[ \prod_{(i,j) \in I} N(V_l^{ij}|f(\alpha_i t_{ij} + \beta_i|\theta_{Z_l}), \sigma_{Z_l}) \prod_{l_2 \in N_l} \Psi(Z_l, Z_{l_2}) \right]$$

where $\Psi(Z_l, Z_{l_2})$ is a clique term representing the likelihood of a neighbouring vertex $l_2$ to have similar label with vertex $l$. The formula for the clique term is:

$$\Psi(Z_l = k, Z_{l_2} = k_2) = \begin{cases} exp(g(\lambda)) & \text{if } k = k_2 \\ exp(-h(\lambda)) & \text{otherwise} \end{cases}$$

where $\lambda$ is a parameter controlling how much to penalise neighbouring vertices that belong to distinct clusters, and $g$ and $h$ are positive, monotonic functions over the $\lambda > 0$ range. We choose $g(\lambda)=\lambda$ and $h(\lambda)=\lambda^2$, which results in a concave objective function for $\lambda$, ensuring that it can later be optimised (see M-step).

Therefore, the model parameters that need to be estimated are $M = [\alpha, \beta, \theta, \sigma, \lambda]$ where $\alpha$ and $\beta$ are the subject specific shifting parameters, $\theta$ and $\sigma$ are the cluster specific trajectory and noise parameters and $\lambda$ is the clique parameter denoting the penalisation of spatially non-smooth assignments of latent variables $Z$.

## 2.2 Fitting the Model using Generalised Expectation-Maximisation

We choose to fit our model using Expectation-Maximisation (EM), because it offers a fast convergence given the large number of parameters that need to be estimated and the huge dimensionality of relevant datasets (e.g. 1973 subjects x 163,842 vertices in ADNI). In the next two sections we outline the E-step and M-step. While both of these steps have no closed-form solution, we will solve them using numerical optimisation, which only results in an increase in the objective function at each iteration. However, the EM algorithm is still guaranteed to converge, and this approach is called Generalised EM (Bishop, 2006).

Algorithm 1 shows the model fitting procedure using the EM algorithm. The procedure first initialises (line 1) some parameters required to start the EM algorithm: the subject time shifts α and β and the latent parameters $z_{lk}$ which represent the assignment of vertices to clusters. In the M-step, the method updates the trajectories of each cluster (lines 4-6), the subjects-specific time shifts (line 9) and the clique penalty term λ (line 17). In the E-step, the method computes $z_{lk}$ (line 18) using previously defined functions that compute $z_{lk}$ given a fixed λ (line 14).

```
1  Initialise α^(0), β^(0), z_lk^(0)
2  while θ, σ, α, β or z_lk not converged do
       ; // M-step 1: For each cluster, optimise its trajectory
3      for k = 1 to K do
4          θ_k^(u) = arg min_{θ_k} Σ_{l=1}^L z_lk^(u-1) Σ_{(i,j)∈I} (V_l^ij − f(α_i^(u-1) t_ij + β_i^(u-1) | θ_k))^2 − log p(θ_k)
5          θ_k^(u) = make_identifiable(θ_k^(u))
6          (σ_k^(u))^2 = (1/|I|) Σ_{l=1}^L z_lk^(u-1) Σ_{(i,j)∈I} (V_l^ij − f(α_i^(u-1) t_ij + β_i^(u-1) | θ_k^(u)))^2 − log p(σ_k)
7      end
       ; // M-step 2: For each subject, optimise its time shift α_i and progression speed β_i
8      for i = 1 to S do
9          α_i^(u), β_i^(u) = arg min_{α_i,β_i} [ Σ_{l=1}^L Σ_{k=1}^K z_lk^(u-1) / (2(σ_k^(u))^2) Σ_{j∈I_i} (V_l^ij − f(α_i t_ij + β_i | θ_k^(u)))^2 ] − log p(α_i, β_i)
10     end
       ; // E-step 1: Define functions ζ_lk(λ) computing z_lk, the probability of vertex l being assigned to cluster k, given fixed λ
11     for l = 1 to L do
12         for k = 1 to K do
               ; // Pre-compute data fit terms D_lk
13             D_lk = −(1/2) log (2π (σ_k^(u))^2) |I| − 1/(2(σ_k^(u))^2) Σ_{i,j∈I} (V_l^ij − f(α_i^(u) t_ij + β_i^(u) | θ_k^(u)))^2
14             ζ_lk(λ) ≈ exp ( D_lk + Σ_{l_2∈N_l} log [ exp(−λ^2) + z_{l_2 k}^(u-1) (exp(λ) − exp(−λ^2)) ] )
15         end
16     end
       ; // M-step 3: optimise clique term λ using above definitions in E-step 1
17     λ^(u) = arg max_λ Σ_{l=1}^L Σ_{k=1}^K ζ_lk(λ) [ D_lk + λ Σ_{l_2∈N_l} ζ_{l_2 k}(λ) − λ^2 Σ_{l_2∈N_l} (1 − ζ_{l_2 k}(λ)) ]
       ; // E-step 2: Compute next z_lk using the best λ
18     z_lk^(u) = ζ_lk(λ^(u))
19 end
20 α_i^(u) = α_i^(u) / σ_N,  β_i^(u) = (β_i^(u) − μ_N) / σ_N  ; // Re-scale subject shifts
```

Algorithm 1. The optimisation procedure for fitting DIVE using the Expectation-Maximisation algorithm.

### 2.2.1 E-step

In the Expectation step, at iteration *u* we seek an estimate of *p(Z / V, M$^{(u-1)}$)*, given the

current estimates of the parameters $M^{(u-1)} = [\theta_k^{(u-1)}, \sigma_k^{(u-1)}, \alpha_i^{(u-1)}, \beta_i^{(u-1)}, \lambda_i^{(u-1)}]$. While, this does not directly factorise over the vertices $l$ due to the MRF terms, we condition the clique terms on the values of $Z$ from the previous iterations. This approximation gives the following factorizable likelihood:

$$p(Z|V, M^{(u-1)}) \approx p(Z|V, M^{(u-1)}, Z^{(u-1)}) = \prod_{l=1}^{L} p(Z_l|V_l, M^{(u-1)}, Z_{N_l}^{(u-1)})$$

The factorised form allows for tractable computation and memory storage of $p(Z)$. Let $z_{lk}^{(u)} = p(Z_l = k \mid V_l, M^{(u-1)}, Z^{(u-1)})$. After simplifications we reach the following update rule:

$$\log z_{lk} \propto \left[ -\frac{\log(2\pi\sigma_k^2)|I|}{2} - \sum_{i,j \in I} \frac{1}{2\sigma_k^2}(V_l^{ij} - f(\alpha_i t_{ij} + \beta_i|\theta_k))^2 \right] +$$

$$\left[ \sum_{l_2 \in N_l} \log \exp(-\lambda^2) + z_{l_2 k}^{(u-1)}(\exp(\lambda) - \exp(-\lambda^2)) \right]$$

The full derivation is given in the Supplementary material. In order to enable optimisation over $\lambda$, a final modification of this step is performed, by considering $z_{lk}$ to be functions $\zeta_{lk}(\lambda)$ over $\lambda$. This results in the update equation from Alg. 1, line 18 which is based on pre-defined terms on lines 13-14.

### 2.2.2 M-step

In the Maximisation step we try to estimate the model parameters $M = (\alpha, \beta, \theta, \sigma, \lambda)$ that maximise $E_{Z|V,M^{(u-1)}}[\log p(V,Z|M)]$. We cannot simultaneously optimise all 5 sets of parameters, so we optimise them independently. In order to get the update rule for the trajectory parameters $\theta_k$ corresponding to cluster $k$ we need to maximise the expected log likelihood with respect to $\theta_k$. The key observation here is that if we assume fixed $\alpha$, $\beta$ and $Z$, then the trajectory parameters $\theta_k$ for every cluster $k$ are conditionally independent, i.e. $\theta_k \perp \theta_m \mid (Z, \alpha, \beta, \sigma) \; \forall \; (k, m), k \neq m$. This allows us to maximise every $\theta_k$ independently using the following equation:

$$\theta_k = \arg\max_{\theta_k} \sum_{z_1,...,z_L}^{K} p(Z = (z_1,...,z_L)|V, M^{(u-1)}) \log \left[ \prod_{l=1}^{L} \prod_{(i,j) \in I} N(V_l^{ij}|f(\alpha_i t_{ij} + \beta_i|\theta_{z_l}), \sigma_{z_l}) \right] + \log p(\theta_k)$$

A similar observation of conditional independence can also be observed for the latent variables $Z$. This allows us to decompose the joint distribution over $Z$, and after expanding the noise model we reach the optimisation problem from Alg. 1, line 4. See Supplementary material for full derivation. This does not have a closed-form solution, so we use numerical optimisation for finding $\theta_k$ that maximises the equation from Alg. 1, line 4.

A similar equation, yet in closed form, is also obtained for $\sigma_k$ (Alg. 1, line 6). After estimating $\theta$ and $\sigma$ for every cluster, we use the new values to estimate the subject

specific parameters α and β. For every subject *i*, we maximise the expected log likelihood with respect to $α_i$, $β_i$ independently, and after simplifications we obtain the update rule from Alg. 1, line 9, which is again solved using numerical optimisation. For the numerical optimisation of θ we used the Nelder-Mead method for its robustness, while for α and β we used the second-order Broyden–Fletcher–Goldfarb–Shanno algorithm due to fast convergence. Finally, we achieved a significant speed-up in the evaluation of objective functions by computing a $z_{lk}$-weighted average of vertex measurements within each cluster (see Supplementary section 4). This resulted in a convergence time of around 6 hours for the larger datasets (ADNI).

For optimising λ, we again try to optimise in the M-step the expected full data likelihood under the *Z* estimates from the previous iteration:

$$\lambda^{(u)} = \arg\max_{\lambda} E_{p(Z|V,M^{(u-1)},\lambda,Z^{(u-1)})}[log\ p(V,Z|M^{(u-1)})]$$

We simplify the above equation by expanding the likelihood model and approximating the joint over *Z* with the product of the marginals $z_{lk}$ over all vertices *l*. This results in the update equation from Alg. 1 line 17 – see supplementary material for full derivation. In this final equation we also replaced $z_{lk}$ with a function $\zeta_{lk}(\lambda)$ over λ, which updates $z_{lk}$ based on the current value of λ being evaluated. This is done to increase convergence, as latent variables $z_{lk}$ are highly coupled with the value of λ being evaluated.

## 2.3 Implementation Details
### 2.3.1 Parameter initialisation and priors

Before starting the fitting process, we need to initialise α, β and the clustering probabilities $z_{lk}$ (Alg. 1, line 1). We set $α_i$ and $β_i$ to be 1 and 0 respectively for each subject, which sets the initial disease progression score to the age of the subject at the clinical visit. We initialise $z_{lk}$ using k-means clustering of the vectors $V_l$. The DPS scores have two extra degrees of freedom (scale and shift), which we account for by setting informative gamma and Gaussian priors on parameters $α_i$ and $β_i$ respectively ($α_i \sim \Gamma(16e4, 16e4)$, $β_i \sim N(0, 0.1)$), which work well in practice as they result in realistic ranges for $α_i$ and $β_i$ of around [0.3, 3] and [-15,15] respectively. Such informative priors on $α_i$ and $β_i$ also help deal with singularities in the objective functions of $α_i$ and $β_i$ when the biomarker trajectories are flat. As already explained in (Jedynak et al., 2012), the sigmoid parameters $θ_k$ are not identifiable, so we need to apply the following transformation on line 5 of Alg. 1: if $b_k^{(u)} < 0$ then $a_k^{(u)} = -a_k^{(u)}$; $b_k^{(u)} = -b_k^{(u)}$; $d_k^{(u)} = d_k^{(u)} - a_k^{(u)}$. This ensures model identifiability and is performed at every iteration.

### 2.3.2 Estimating the Optimal Number of Clusters

The EM procedure needs to specify a-priori the number of clusters to fit on the data. We optimise the number of clusters $K$ using Akaike Information Criterion (AIC), which we found to better agree with ground truth in simulations than other information criteria such as the Bayesian Information Criteria (BIC). The number of parameters of the fitted model is $5K+2S+1$, where $S$ is the number of subjects. Note that $z_{lk}$ are not included as parameters of the model because they are latent variables that are marginalised (see Eq. 6). We repeat the fitting procedure for each $K$ from 2 to 100 clusters and select the $K$ that minimises the AIC.

## 2.4 Simulation Experiments

### 2.4.1 Motivation

Initial assessment of DIVE performance uses synthetic data, where we know the ground truth. The aim is to explore how accurately we can recover ground truth parameters as the problem becomes harder in three different scenarios:
Scenario 1: as the number of clusters increases, evaluate how well DIVE can estimate the correct number of clusters using AIC and BIC
Scenario 2: as the trajectories become more similar, test how well we can recover the assignment of vertices to clusters and the DIVE parameters
Scenario 3: same as Scenario 2, but for decreasing number of subjects

### 2.4.2 Synthetic Data Generation

We first designed a basic simulation, which the model should be able to fit well since the trajectories were designed to be well separated and enough subject data was generated along the disease time course. The data in the basic simulation was generated as follows:
1. sample age $a_{i1}$ and time shift parameters $\alpha_i$, $\beta_i$ for 300 subjects with 4 timepoints (each timepoint 1 year apart), with $a_{i1} \sim U(40,80)$, $\alpha_i \sim \Gamma(6.25, 6.25)$, $\beta_i \sim N(0, 10)$. Time since baseline has been obtained for every visit $j$ of subject $i$ as follows: $t_{ij} = a_{ij} - a_{i1}$
2. generate three sigmoids with different (slope, centre) parameters: [(-0.1, -15), (-0.1, 2.5), (-0.1, 20)] (Fig. 2A, red lines). Upper and lower limits have been set to 1 and 0 respectively.
3. randomly assign every vertex $l \in \{1, ..., L\}$ where $L = 1000$, to a cluster $a[l] \in \{1,2,3\}$
4. sample a set of $L$ perturbed trajectories $\theta_l$ from each of the original trajectories, one for each vertex (Fig. 2A, gray lines) using covariance matrix $C_\theta = diag([0, 2b_k/15, 11.6, 0])$.
5. sample subject data for every vertex $l$ from its corresponding perturbed trajectory $\theta_l$ with noise standard deviation $\sigma_l = 1$.

From the basic simulation, we generated synthetic data for each of the three scenarios by varying one parameter at a time and kept the other parameters constant, having the same values as in the basic simulation. We varied the following parameters:
Scenario 1: number of clusters - 2, 3, 5, 10, 15, 20, 30 and 40. The cluster centres were spread evenly across a fixed total DPS range where data was available.
Scenario 2: distance between trajectory centres (as proportion of total DPS range sampled) – 0.33, 0.30, 0.23, 0.17, 0.10, 0.07, 0.03 and 0.02
Scenario 3: number of subjects - 300, 200, 100, 50, 35, 20, 10 and 5

### 2.4.3 Model Fitting and Evaluation

Since there was no spatial information in the data generation procedure, we used DIVE without the MRF extension. For Scenario 1, we estimated using AIC and BIC the optimal number of clusters. For Scenarios 2 and 3, after fitting the parameters of DIVE, we calculated the agreement between the final clustering probabilities $p(Z_l)$ and the true clustering assignments $a[l]$. This agreement, which we will call the *clustering agreement,* is defined as $\aleph = max_\tau (1/L)\Sigma^L_{l=1} p(Z_l = \tau(a[l]))$, *where τ is any permutation of cluster labels*. We also computed the error in the DPS estimation (sum of squared differences, SSD) and trajectory estimation (SSD between predicted trajectory and true trajectory at DPS points of every subject visit).

## 2.5 Data Acquisition and Pre-processing

Data used in this work were obtained from the Alzheimer's Disease Neuroimaging Initiative (ADNI) database (adni.loni.usc.edu) and from the Dementia Research Centre, UK. For ADNI, we downloaded all T1 MR images that have undergone gradient warping, intensity correction, and scaling for gradient drift. We included subjects that had at least 3 scans, to ensure we get a robust estimate of the subject specific parameters. This resulted in 138 healthy controls, 235 subjects with mild cognitive impairment (MCI) and 81 subjects with Alzheimer's disease.

We also downloaded all AV45 PET images from ADNI that were fully pre-processed, having the tag *'Co-reg, Avg, Std Img and Vox Siz, Uniform Resolution'*. This meant that the images were co-registered, averaged across the 6 five-minute frames, standardised with respect to the orientation and voxel size and smoothed to produce a uniform resolution of 8mm full-width/half-max (FWHM).

The DRC dataset consisted of T1 MRI scans from 31 healthy controls, 32 PCA and 23 typical AD subjects with at least 3 scans each and an average of 5.26 scans per subject. All PCA patients fulfilled the (Tang-Wai et al., 2004) criteria and (Mendez et al., 2002) criteria based on clinical review. The typical AD patients all met the criteria for probable Alzheimer's disease (Dubois et al., 2010; Dubois et al., 2007).

Given that the ADNI and DRC datasets contained subjects with different modalities

or diseases, we ran DIVE independently on the following four cohorts (see Table 1 for demographics):
1) *ADNI MRI*: controls, MCI and tAD subjects from ADNI (cortical thickness data)
2) *DRC tAD*: tAD subjects and controls from the DRC dataset (cortical thickness data)
3) *DRC PCA*: PCA subjects and controls from the DRC dataset (cortical thickness data)
4) *ADNI PET*: AV45 scans from ADNI containing subjects with following diagnoses: healthy controls, subjective memory complaints, early MCI, late MCI and Alzheimer's disease.

| Cohort | Diagnosis | Number of Subjects | Number of Scans | Age at baseline (years) |
|---|---|---|---|---|
| ADNI MRI | Controls | 138 | 4.3 | 76.3 |
| | MCI | 235 | 4.6 | 74.8 |
| | AD | 81 | 3.5 | 75.8 |
| DRC tAD | Controls | 31 | 5.0 | 66.3 |
| | AD | 24 | 5.4 | 71.2 |
| DRC PCA | Controls | 31 | 5.0 | 66.3 |
| | PCA | 32 | 4.1 | 62.6 |
| ADNI PET | Controls | 141 | 2.4 | 85.5 |
| | SMC | 27 | 2.0 | 86.1 |
| | EMCI | 149 | 2.4 | 85.6 |
| | LMCI | 104 | 2.4 | 86.0 |
| | AD | 12 | 2.0 | 87.3 |

Table 1. Demographics of the three cohorts used in our analysis. ADNI MRI and the DRC cohorts were used for the cortical thickness analysis, while ADNI PET was used for the PET AV45 analysis. MCI – mild cognitive impairment, SMC - subjective memory complaints, EMCI – early MCI, LMCI – late MCI.

### 2.5.1 Image Pre-processing

*2.5.1.1 MR Images*

On both datasets, in order to extract reliable cortical thickness measures, we ran the Freesurfer longitudinal pipeline (Reuter et al., 2012), which first registers the MR scans to an unbiased within-subject template space using inverse-consistent registration. The longitudinally registered images were then registered to the average Freesurfer template. No further smoothing was performed on these images (FWHM level of zero mm). From these template-registered volumetric images, cortical thickness measurements were computed at each vertex (i.e. point) on an average 2D cortical surface manifold. For each vertex we averaged the thickness levels from both hemispheres. Finally, we standardised the data from each vertex with respect to the values of that vertex in the control population. Each of the final images had a resolution of 163,842 vertices on the cortical surface.

*2.5.1.2 PET AV45 Images*

We computed amyloid SUVR levels using the PetSurfer pipeline (Greve et al., 2016; Greve et al., 2014), which is available with Freesurfer version 6. The PetSurfer pipeline first registers the PET image with the corresponding MRI scan, then applies Partial Volume Correction, and finally resamples the voxelwise SUVR values onto the cortical surface. While the final images also had a resolution of 163,842 vertices, the PET data we obtained from ADNI was inherently more smooth than the MRI cortical thickness data (8mm FWHM).

### 2.5.2 The MRF Neighbourhood Graph

We estimated the MRF neighbourhood graph based on a Freesurfer triangular mesh for the *fsaverage* template. Each vertex was a triangle on the brain surface estimated with Freesurfer, and we connected the vertices if the corresponding triangles had a shared edge. For the MRF neighbourhood graph, we used a $3^{rd}$ degree neighbourhood structure, meaning that two vertices were considered neighbours if the shortest path between them was not higher than 3.

## 3 Results

We first present results on synthetic data (section 3.1), then on ADNI and DRC datasets (section 3.2), followed by model evaluation (section 3.3) using cross-validation and correlation with cognitive markers.

### 3.1 Results on synthetic data

In the basic simulation, we obtained a clustering agreement ℵ of *0.97*, which suggests that almost all vertices were assigned to the correct cluster. Figure 2A shows the original trajectories and the recovered trajectories using our model, plotted against the disease progression score on the x-axis and the vertex value on the y-axis. In Fig. 2B we plotted the recovered DPS of each subject along with the true DPS. The results for the three scenarios are shown in Figs. 2C-E. In Fig 2C, we show for Scenario 1 the estimated number of clusters against the true number of clusters using both AIC and BIC criteria. In Figs 2D-E we show the distributions for ℵ in Scenarios 2 and 3 as the problem becomes harder in each successive step.

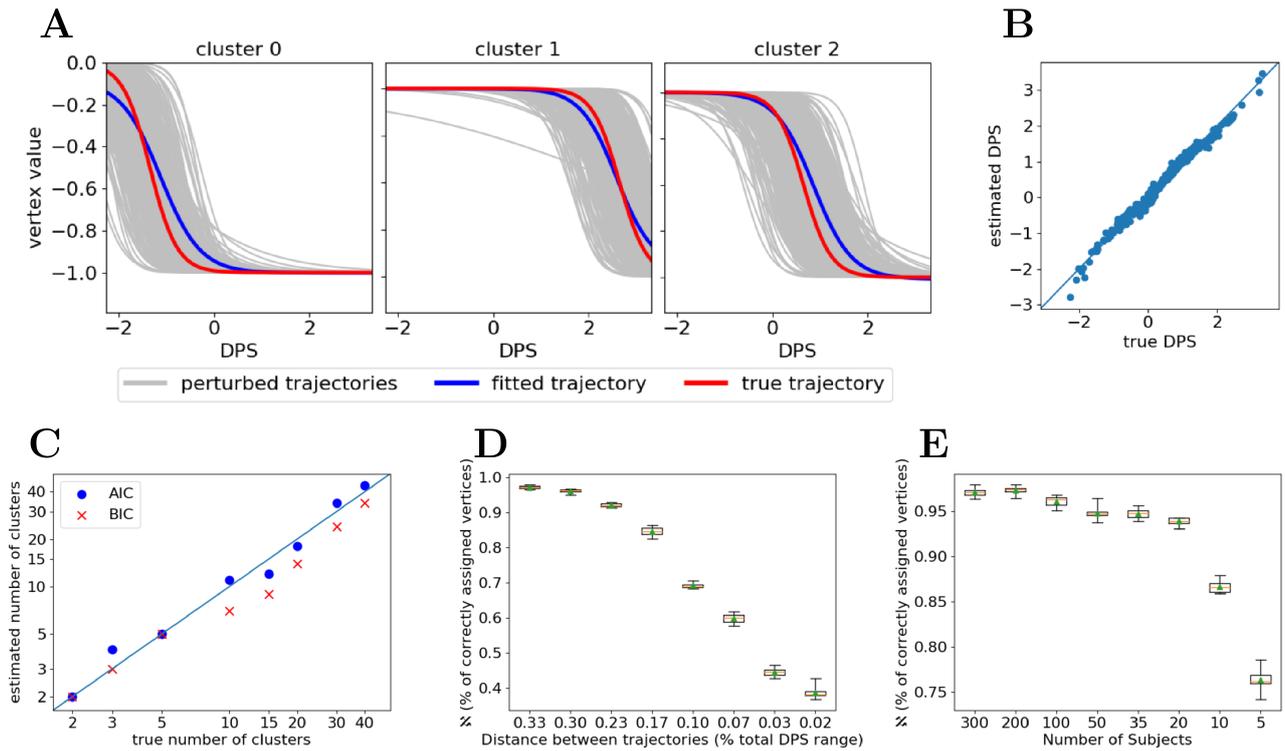

Figure 2. (A-B) Results for the basic simulation, where trajectories are relatively well separated. (A) Reconstructed temporal trajectories (blue) plotted against the true trajectories (red). The x-axis shows the disease progression score (DPS), while the y-axis shows the biomarker values of the vertices. (B) Estimated subject-specific DPS scores compared to the true scores. (C-E) Simulation results for the three scenarios: (C) increasing number of clusters, (D) trajectories becoming similar and (E) decreasing number of subjects. On the x-axis we show the variable that was changing within the scenario (e.g. number of clusters), while on the y-axis we show the agreement measure ℵ, representing the percentage of vertices that were assigned to the correct cluster.

The results show that, in a simple experiment where the trajectories are well separated, DIVE can very accurately estimate which clusters generated each vertex. Moreover, the recovered trajectories and DPS scores are close to the true values. The results of Scenario 1 also suggest that both AIC and BIC are effective at estimating the correct number of known clusters, with AIC having slightly better performance than BIC for larger numbers of clusters. On the other hand, the results of the stress test scenarios 2 and 3 show that performance measure ℵ drops when the trajectories become very similar with each other or when the number of subjects decreases. This happens because small differences in trajectories are hard to detect in the presence of measurement noise, while a small number of subjects doesn't provide enough data to accurately estimate the parameters. Similar decreases in performance for scenarios 2 and 3 are observed also for other measures, such as the error in recovered trajectories or DPS scores (Supplementary Fig 1).

## 3.2 Results on ADNI and DRC Datasets

### 3.2.1 Initial Hypotheses

Using ADNI and DRC datasets, we aim to recover the spatial distribution of cortical atrophy and amyloid pathology, as well as the rate and timing of these pathological processes. In particular, we hypothesize that these spatial patterns of pathology and their evolution will be:
1. similar on two independent typical AD datasets: ADNI and DRC
2. different on distinct diseases: tAD vs PCA
3. different in distinct modalities: cortical thickness from MRI vs amyloid load from AV45 PET.

### 3.2.2 Results

The optimal number of clusters, as estimated with AIC, was three for the ADNI MRI dataset, three for the DRC tAD dataset, five for the DRC PCA dataset and eighteen for the ADNI PET dataset. Fig. 3A-left shows the results from the ADNI MRI dataset, where in the left image we coloured the vertices on the cortical surface according to the cluster they most likely belong to. We assigned a colour for each cluster (both the brain figures on the left and the trajectory figures on the right) according to the extent of pathology of its corresponding trajectory at a DPS score of 1. The cluster colours range from red (severe pathology) to blue (moderate pathology). In Fig. 3A-right, we show the resulting cluster trajectories with samples from the posterior distribution of each $\theta_k$. Similar results are shown for the other three datasets: the DRC tAD dataset (Fig. 3B), DRC PCA dataset (Fig. 3C) and the ADNI PET dataset (Fig. 3D).

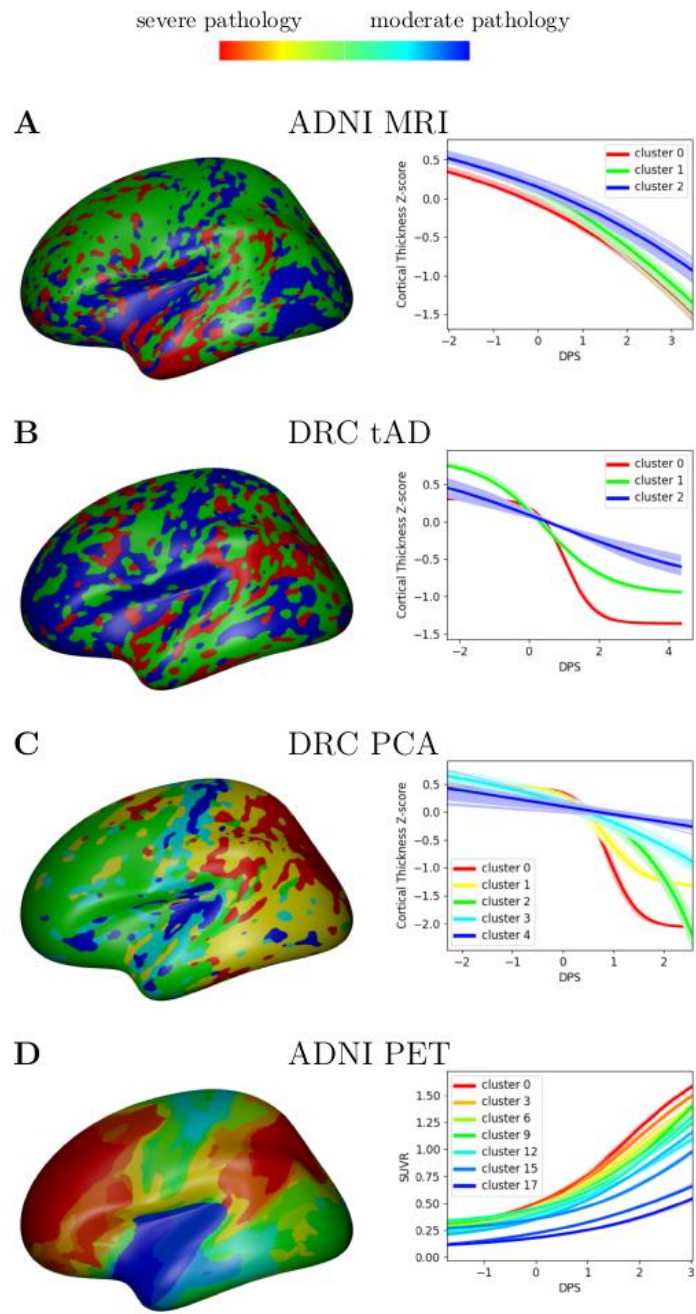

Figure 3. (left column) DIVE estimated clusters (left column) and corresponding disease progression trajectories (right column) on four datasets: (A) ADNI MRI (B) DRC tAD (C) DRC PCA and (D) ADNI PET. We coloured each cluster according to the extent of pathology (cortical thickness or amyloid uptake) at DPS=1.

We notice that in tAD subjects using the ADNI datasets (Fig. 3A), there is more severe cortical thinning mainly in the inferior temporal lobe (red cluster), with disperse atrophy also in parietal and frontal regions (green cluster), with relative sparing of the inferior frontal and occipital lobes. In tAD subjects from the DRC dataset, we see a relatively similar pattern, however with more pronounced atrophy in the supramarginal cortex (red cluster) compared to ADNI. The spatial distribution of cortical thinning found with DIVE resembles results from previous longitudinal studies such as (Dickerson et al., 2008; Singh et al., 2006). However, in contrast to

these approaches, our model gives insight into the timing and rate of atrophy and is also able to stage subjects across the disease time course. We also find that the cluster trajectories in the DRC tAD dataset have similar dynamics to the ADNI MRI dataset, although they show a clearer separation between each other.

In the PCA subjects (Fig. 3C), we find that atrophy is mainly focused on the posterior part of the brain, mostly the posterior parietal and supramarginal areas, with limited spread in the motor cortex, anterior temporal and frontal areas. There is a clear separation of atrophy extent between the anterior and posterior part of the brain, with the boundary running along the motor cortex. This posterior-focused pattern of atrophy is different from the one found in the tAD datasets, and agrees with previous findings in the literature (Crutch et al., 2012; Lehmann et al., 2011).

In ADNI PET (Fig. 3D) we see that the regions with the highest amyloid uptake are more spatially continuous, comprising the precuneus and anterior frontal areas. On the other hand, the anterior-superior temporal gyrus shows the least uptake of amyloid. This result closely matches the result by (Bilgel et al., 2016), which used a completely different dataset and modelling technique. The "layers of clusters" starting from the precuneus and frontal lobes, which range from severe to less severe atrophy, suggest a continuum of variation in vertex trajectories in the case of the PET dataset (Fig 3D-right). These trajectories all start with a low amyloid SUVR, between 0 and 0.25, but in late stages the trajectories for some clusters such as cluster 0 can reach an SUVR of 1.5. The reason for seeing this continuum might be because the PET images have a much lower resolution than MR images and were smoothed by ADNI during the pre-processing steps.

## 3.3 Model Evaluation
### 3.3.1 Motivation

We further tested the robustness and validity of the model as follows:
1. *Robustness in parameter estimation*: test whether similar spatial clustering is estimated for different subsets of the data
2. *Clinical validity of DPS scores*: test whether the subject disease progression scores, based purely on MRI or PET data, correlate with cognitive tests such as Clinical Dementia Rating Scale - Sum of Boxes (CDRSOB), Alzheimer's Disease Assessment Scale - Cognitive (ADAS-COG), Mini-Mental State Examination (MMSE) and Rey Auditory and Verbal Learning Test (RAVLT).
3. *Prediction accuracy*: given initial baseline scans from different subjects, test how accurately DIVE can predict the biomarker evolution of those patients over time.
4. *Comparison with other models*: we compared the performance of DIVE with a region-of-interest based method (Jedynak et al., 2012) and a no-staging method that doesn't estimate subject time shifts. See Supplementary Section 2 for precise specifications.

### 3.3.2 Evaluation Procedure

For all scenarios, we ran 10-fold cross-validation (CV) on the ADNI MRI dataset. At each fold we fit the model using 3 clusters, since this was the optimal number of clusters found previously on the entire dataset. The trained model was then used to estimate the DPS of the test subjects.

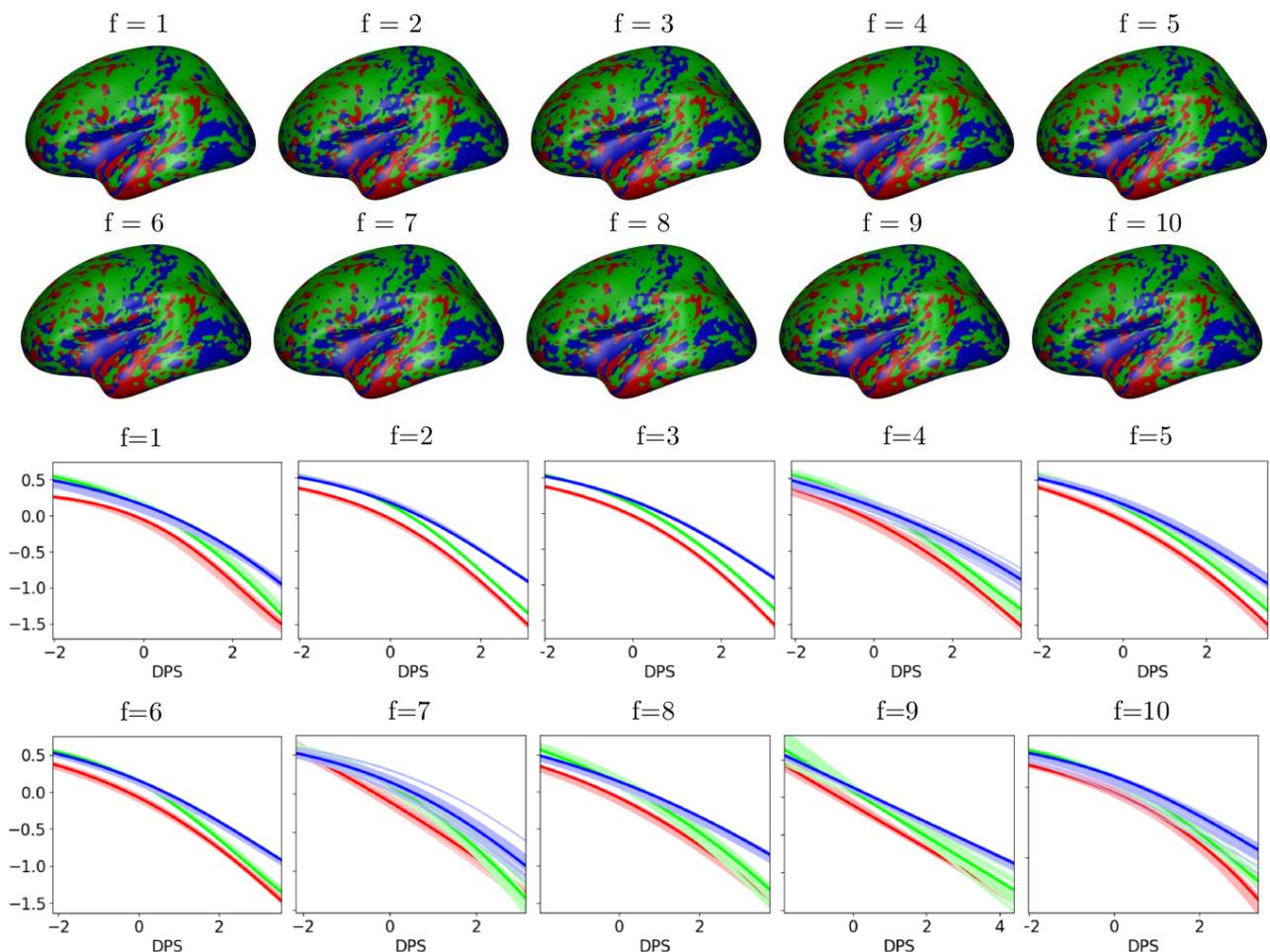

Figure 4. (top) Clusters estimated from 10-fold cross-validation training sets on the ADNI MRI dataset. (bottom) Estimated trajectories for each fold.

### 3.3.3 Evaluation Results

Fig. 4 shows the brain clusters and corresponding trajectories, estimated for all the cross-validation folds after fitting the model on the training data. The clusters have been coloured using a similar colour scheme as in Fig. 3. In Fig 5 we show scatter plots of the DPS scores with clinical measures such as CDRSOB, ADAS-COG, MMSE and RAVLT.

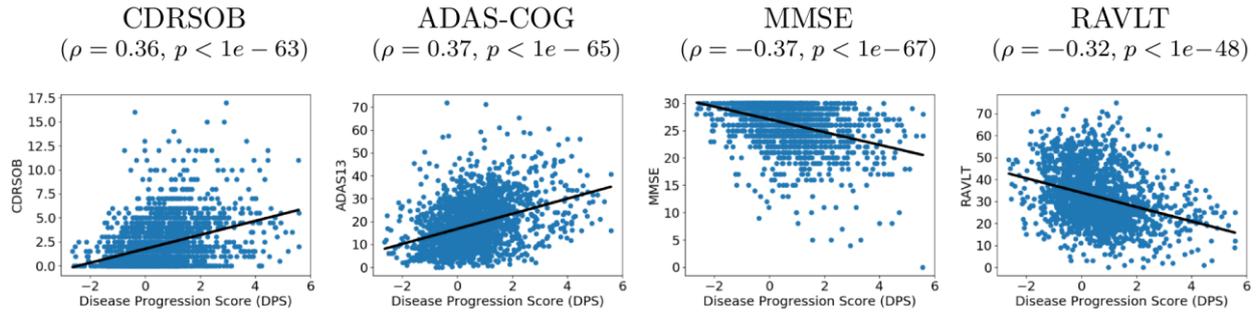

Fig 5. Scatter plots of the DPS scores estimated from the ADNI MRI dataset, plotted against four cognitive tests: CDRSOB, ADAS-COG, MMSE and RAVLT. For each cognitive test we also report the correlation coefficient and p-value. The disease progression scores, computed only based on MRI cortical thickness data, correlate with these cognitive measures, suggesting that the DPS scores are clinically meaningful.

| Model | CDRSOB ($\rho$) | ADAS13 ($\rho$) | MMSE ($\rho$) | RAVLT ($\rho$) | Prediction (RMSE) |
|---|---|---|---|---|---|
| DIVE | 0.37 +/- 0.09 | 0.37 +/- 0.10 | 0.36 +/- 0.11 | 0.32 +/- 0.12 | 1.021 +/- 0.008 |
| ROI-based model | 0.36 +/- 0.10 | 0.35 +/- 0.11 | 0.34 +/- 0.13 | 0.30 +/- 0.13 | 1.019 +/- 0.010 |
| No-staging model | *0.09 +/- 0.06 | *0.03 +/- 0.09 | *0.05 +/- 0.06 | *0.02 +/- 0.06 | *1.062 +/- 0.024 |

Table 2. Performance evaluation of DIVE and two other models on the ADNI MRI dataset using 10-fold cross-validation. In the middle four columns, we show the correlation of the DPS scores with respect to several cognitive tests: CDRSOB, ADAS-Cog13, MMSE and RAVLT. The last column shows the prediction error (RMSE) of cortical thickness values from follow-up scans. (*) Statistically significant differences between the model and DIVE, Bonferroni corrected for multiple comparisons.

The results in Fig. 4 demonstrate that DIVE is robust in cross-validation, as the estimated clusters and trajectory parameters are all similar across folds. The average Dice score overlap across the 10-folds range were 0.77, 0.76 and 0.90 for clusters 0, 1 and 2 respectively. The DIVE-derived DPS scores, which were estimated purely based on MRI data, are also clinically relevant as they correlate with cognitive tests (Fig 5).

The performance of DIVE in terms of subject staging and biomarker prediction also compares favourably with simpler *no-staging* and *ROI-based* models (Table 2). Results show that DIVE has comparable performance to the *ROI-based* model, both in terms of subject staging and cortical thickness prediction. The fact that DIVE has similar performance to a simpler model which has less parameters is evidence that the estimated patterns are meaningful. Moreover, DIVE offers qualitative insight into the fine-grained spatial patterns of pathology and their temporal progression. Furthermore, the *No-staging model* performs significantly worse than DIVE, both in terms of subject staging and for biomarker prediction. This suggests that, when modelling progression of AD, it is important to account for the fact that patients are

at different stages along the disease time-course.

# 4 Discussion

### 4.1.1 Summary and Key Findings

We presented DIVE, a spatiotemporal model of disease progression that clusters vertex- or voxel-wise measures of pathology in the brain based on similar temporal dynamics. The model highlights, for the first time, groups of cortical vertices that exhibit a similar temporal trajectory over the population. The model also estimates the temporal shift and progression speed for every subject. We applied the model on cortical thickness vertex-wise data from three MRI datasets (ADNI, DRC tAD and DRC PCA), as well as an amyloid PET dataset (ADNI). Our model found qualitatively similar patterns of cortical thinning in tAD subjects using the two independent datasets (ADNI and DRC). Moreover, it also found different patterns of pathology dynamics on two distinct diseases (tAD and PCA) and on different types of data (PET and MRI-derived cortical thickness). Finally, DIVE also provides a new way to parcellate the brain that is specific to the temporal trajectory of a particular disease.

### 4.1.2 Limitations and future work

DIVE has some limitations that can be addressed. First, we assumed that cluster trajectories follow sigmoidal shapes, which is not the case for many types of biomarkers in ADNI which do not plateau in later stages. The assumption of sigmoidal trajectories can be avoided using non-parametric curves such as Gaussian Processes (Lorenzi et al., 2017), which would be straightforward to incorporate into the DIVE framework. Another limitation of the model is that it assumes all subjects follow the same disease progression pattern, which might not be the case in heterogeneous datasets such as ADNI or DRC. This can be a concern, as there might be a pattern of pathology that occurs in a small set of subjects. However, DIVE can be extended to account for heterogeneity in the datasets by modelling different progression dynamics for distinct subgroups, using unsupervised learning methods like the SuStaIn model by (Young et al., 2018). While SuStaIn, just like DIVE, estimates clusters and trajectories within the dataset, the clusters in SuStaIn are made of *subjects* with similar disease progression, while the clusters in DIVE are made of *vertices* with similar progression. Future work could combine clustering along both subjects and vertices simultaneously to estimate disease subtypes with distinct spatiotemporal dynamics at the vertexwise level.

There are several potential future applications of DIVE. One of the advantages of DIVE is that it can be used to study the link between disconnected patterns of brain pathology and connectomes extracted from diffusion tractography or functional MRI (fMRI). Such an analysis would enable further understanding of the exact underlying mechanisms by which the brain is affected by the disease. Our model, which can

estimate fine-grained spatial patterns of pathology, is more suitable than standard ROI-based methods for studying the link between pathology and these structural or functional connectomes, because white matter or functional connections have a fine-grained and spatially-varying distribution of endpoints on the cortex.

Apart from studying the link with brain connectomes, there are other potential applications for DIVE. The model can be applied to study other types of vertexwise or voxelwise data, such as FDG PET, tau PET, DTI or Jacobian compression maps from MRI. Moreover, the model can also be extended to cluster points on the brain surface according to a more complex disease signature, that can be made of two or more biomarkers. For example, using our cortical thickness and amyloid PET datasets from ADNI, we could have clustered points on the brain based on both modalities simultaneously. Such complex disease signatures can offer important insights into the relationships between different modalities and underlying disease mechanisms.

DIVE is a spatiotemporal model that can be used for accurately predicting and staging patients across the progression timeline of neurodegenerative diseases. The spatial patterns of pathology can also be used to test mechanistic hypotheses which consider AD as a network vulnerability disorder. All these avenues can help towards disease understanding, patient prognosis, as well as clinical-trials for assessing efficacy of a putative treatment for slowing down cognitive decline.

## 4.2 Acknowledgements



\maketitle


%

\definecolor{blue3}{HTML}{86B7FC} 
\definecolor{blue1}{HTML}{B5F1FF} 
\definecolor{blue2}{HTML}{E0F9FF} 

\def\ci{\perp\!\!\!\perp}

\newcommand{\fldSynth}{.}

\section{Simulations - Error in Estimated Trajectories and DPS}

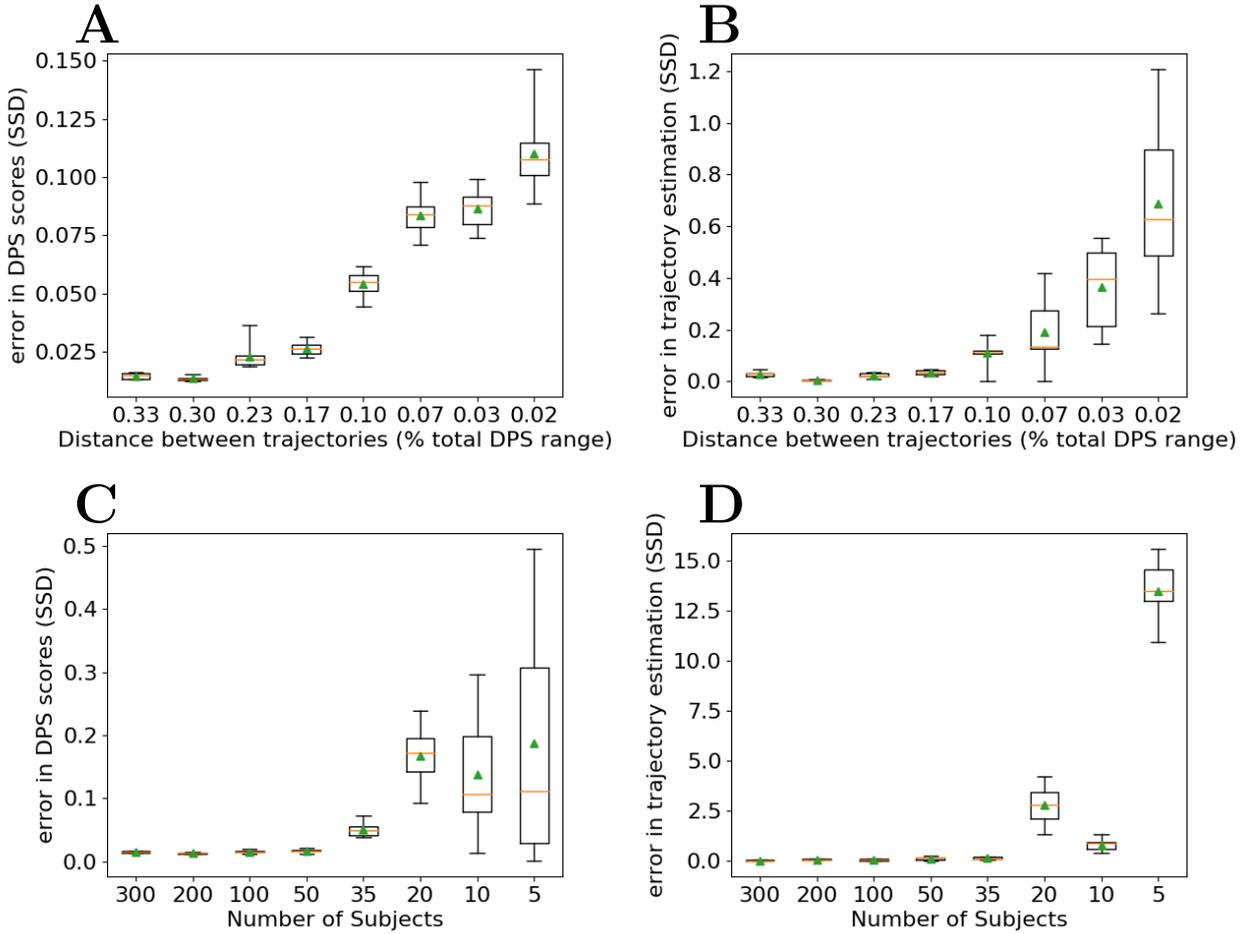
\begin{figure}[H]
\begin{picture}(230,180)
\put(0,0){\includegraphics[width=0.45\textwidth]{\fldSynth/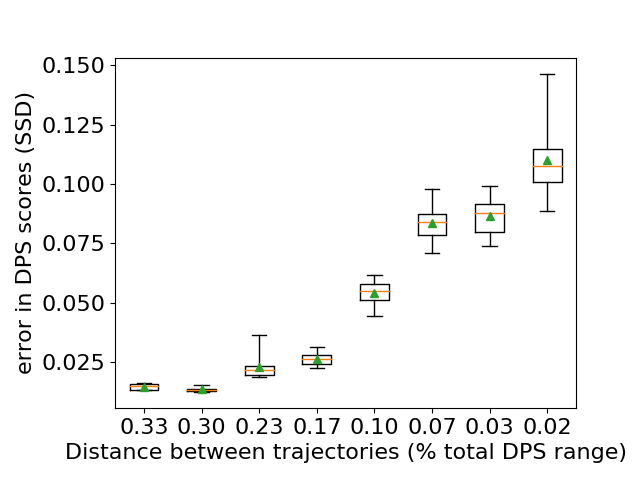}}
\put(30,160){\textbf{\huge{A}}}
\end{picture} 
\begin{picture}(230,180)
\put(0,0){\includegraphics[width=0.45\textwidth]{\fldSynth/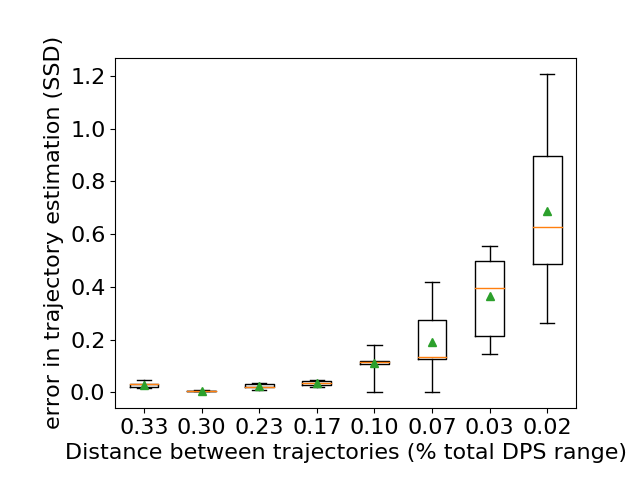}}
\put(30,160){\textbf{\huge{B}}}
\end{picture} 

\begin{picture}(230,180)
\put(0,0){\includegraphics[width=0.45\textwidth]{\fldSynth/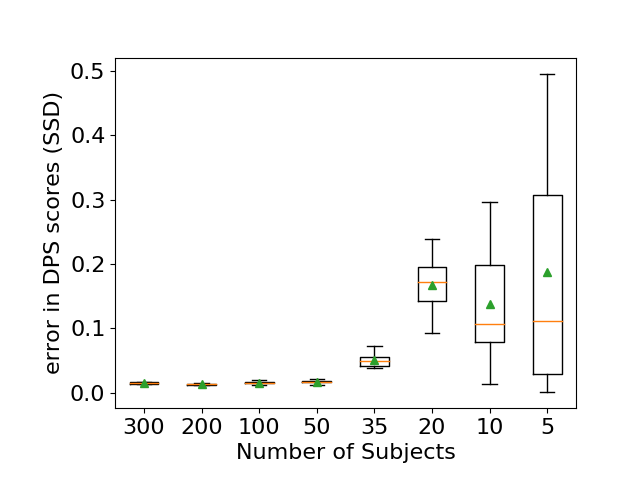}}
\put(30,160){\textbf{\huge{C}}}
\end{picture} 
\begin{picture}(230,180)
\put(0,0){\includegraphics[width=0.45\textwidth]{\fldSynth/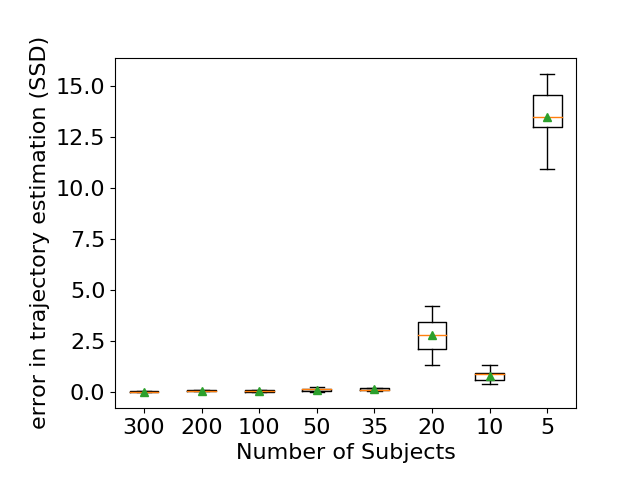}}
\put(30,160){\textbf{\huge{D}}}
\end{picture} 
\caption{Error in DPS scores (A) and trajectory estimation (B) for Scenario 2 in simulation experiments. (C-D) The same error scores for Scenario 3. We notice that as the problem becomes more difficult, the errors in the DIVE estimated parameters increase. Errors were measured as sum of squared differences (SSD) between the true parameters and estimated parameters. For the trajectories, the SSD was calculated only based on the sigmoid centres, due to different scaling of the other sigmoidal parameters.}

\end{figure}

\section{Comparison between DIVE and other models}
 
\subsection{Motivation}

We were also interested to compare the performance of DIVE with other disease progression models. In particular, we were interested to test whether:
\begin{itemize}
 \item Modelling dynamic clusters on the brain surface improves subject staging and biomarker prediction
 \item Modelling subject-specific stages with a linear transformation (the $\alpha_i$ and $\beta_i$ terms) improves biomarker prediction
\end{itemize}

\subsection{Experiment design}

We compared the performance of our model to two simplified models:
\begin{itemize}
 \item ROI-based model: groups vertices according to an a-priori defined ROI atlas. This model is equivalent to the model by Jedynak et al., Neuroimage, 2012 and is a special case of our model, where the latent variables $z_{lk}$ are fixed instead of being marginalised as in equation 6.
 \item No-staging model: This is a model that doesn't perform any time-shift of patients along the disease progression timeline. It fixes $\alpha_i=1$, $\beta_i=0$ for every subject, which means that the disease progression score of every subject is age.
\end{itemize}

We performed this comparison using 10-fold cross-validation. For each subject in the test set, we computed their DPS score and correlated all the DPS values with the same four cognitive tests used previously. We also tested how well the models can predict the future vertex-wise measurements as follows: for every subject i in the test set, we used their first two scans to estimate $\alpha_i=1$, $\beta_i=0$ and then used the rest of the scans to compute the prediction error. For one vertex location on the cortical surface, the prediction error was computed as the root mean squared error (RMSE) between its predicted measure and the actual measure. This was then averaged across all subjects and visits.

\subsection{Results}

Table \ref{tab1} shows the results of the model comparison, on ADNI MRI dataset. Each row represents one model tested, while each column represents a different performance measure: correlations with four different cognitive tests and accuracy in the prediction of future vertexwise measurements. In each entry, we give the mean and standard deviation of the correlation coefficients or RMSE across the 10 cross-validation folds.  

\begin{table}[H]
\centering
\begin{footnotesize}
 \begin{tabular}{c | c c c c | c}
  Model & CDRSOB ($\rho$) & ADAS13 ($\rho$) & MMSE ($\rho$) & RAVLT ($\rho$) & Prediction (RMSE)\\
  \hline 
DIVE & 0.37 +/- 0.09 & 0.37 +/- 0.10 & 0.36 +/- 0.11 & 0.32 +/- 0.12 & 1.021 +/- 0.008 \\
ROI-based model & 0.36 +/- 0.10 & 0.35 +/- 0.11 & 0.34 +/- 0.13 & 0.30 +/- 0.13 & 1.019 +/- 0.010 \\
No-staging model & *0.09 +/- 0.06 & *0.03 +/- 0.09 & *0.05 +/- 0.06 & *0.02 +/- 0.06 & *1.062 +/- 0.024 \\

 \end{tabular}
 \end{footnotesize}
 \caption{Comparison of our model with two more simplistic models on the ADNI MRI dataset. For each of the three models, we show the correlation of the disease progression scores (DPS) with respect to several cognitive tests: CDRSOB, ADAS13, MMSE and RAVLT. The correlation numbers represent the mean correlation across the 10 cross-validation folds. }
 \label{tab1}
\end{table}


\section{Validation of subject parameters against APOE}

\newcommand{\adniThFld}{.}

\begin{figure}[H]
\includegraphics[width=\textwidth]{\adniThFld/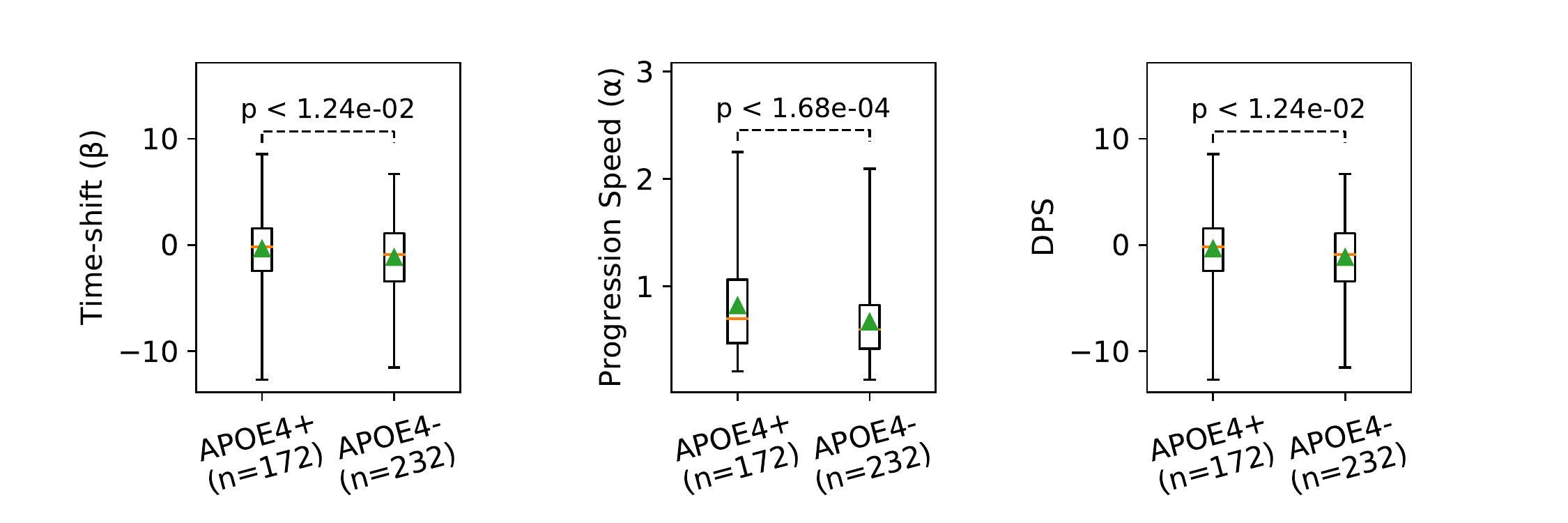}
\caption{Subject-specific parameters (left: time shift parameters $\beta$, middle: progression speed $\alpha$ and right: DPS scores at baseline visit) for APOE4-positive and negative individuals. APOE4 positive subjects have significantly higher time-shifts, progression speeds and overall DPS scores than APOE4-negative individuals.} 
\end{figure}

\newcommand{\Mu}{M}

\section{Derivation of the Generalised EM algorithm}

We seek to calculate $M^{(u)} = \argmax_{M} \mathbb{E}_{Z|V,M^{(u-1)}} \left[log\ p(V,Z|M)\right] + log\ p(M)$  where $M^{(u)} = (\alpha^{(u)}, \beta^{(u)}, \theta^{(u)}, \sigma^{(u)}, \lambda^{(u)})$ are the set of model parameters at iteration $u$ of the EM algorithm, and $Z = (Z_1, \dots, Z_L)$ is the set of discrete latent variables, where $Z_l$ represents the cluster that voxel $l$ was assigned to, so $Z_l \in {1, \dots, K}$. While $Z_l$ (with capital letter) is a random variable, we will also use the notation $z_l$ (small letter) to represent the value that the random variable $Z_l$ was instantiated to. Finally, $p(M^{(u)})$ is a prior on these parameters that is chosen by the user. Expanding the expected value, we get:

\begin{equation}
\label{eq:EM1}
M^{(u)} = \argmax_{M} \sum_{z1,\dots, z_L}^K p(Z = (z_1, ..., z_L) | V, \Mu^{(u-1)}) \left[log\ p(V,Z|M)\right] + log\ p(M) 
\end{equation}

The E-step involves computing $p\left(Z = (z_1, ..., z_L)| V, \Mu^{(u-1)}\right)$, while the M-step comprises of solving the above equation.

\subsection{E-step}

In this step we need to estimate $p(Z | V, \Mu^{(u-1)})$. For notational simplificy we will drop the $(u-1)$ superscript from $\Mu$

\begin{equation}
 p(Z | V, \Mu) = \frac{1}{C}  p(V, Z | \Mu) =  \prod_l^L \left[ \prod_{(i,j) \in I} N(V_l^{ij} | f(\alpha_i t_{ij} + \beta_i | \theta_{Z_l}), \sigma_{Z_l})  \prod_{l_2 \in N_l} \Psi (Z_{l}, Z_{l_2}) \right]
\end{equation}
where $N_l$ is the set of neighbours of vertex $l$. However, this doesn't directly factorise over the vertices $l$ due to the MRF terms $\Psi (Z_{l}, Z_{l_2})$. It is however necessary to find a form that factorises over the vertices, otherwise we won't be able to represent in memory the joint distribution over all $Z$ variables. If we make the approximation $p(Z | V, \Mu) \approx \prod_l^L p(V_l|Z_l, \Mu)$ then we loose out all the MRF terms and the model won't account for spatial correlation. We instead do a first-degree approximation by conditioning on the values of $Z_{N_l}^{(u-1)}$, the labels of nearby vertices from the previous iteration. The approximation is thus:

\begin{equation}
\label{eq:e_approx}
 p(Z | V, \Mu) \approx \prod_l^L \mathbb{E}_{Z_{N_l}^{(u-1)}|V_l, M} \left[ p(Z_l|V_l, \Mu, Z_{N_l}^{(u-1)}) \right]
\end{equation}

This form allows us to factorise over all the vertices to get $p(Z_l | V_l, \Mu)$:

\begin{equation}
 p(Z_l | V_l, \Mu) \approx \frac{1}{C} \sum_{Z_{N_l}^{(u-1)}} p(V_l|Z_l, \Mu) p(Z_l|Z_{N_l}^{(u-1)}) p(Z_{N_l}^{(u-1)}|V_l, M) 
\end{equation}

where $C$ is a normalistion constant that can be dropped. We can now  further factorise $p(Z_l|Z_{N_l}^{(u-1)}) \approx  \prod_{m \in \{1, ..., N_l\}} p(Z_l | \Mu,  Z_{N_l(m)}^{(u-1)} = z_{N_l(m)})$ and apply a similar factorisation to the prior $p(Z_{N_l}^{(u-1)}|V_l, M) $, resulting in:

\begin{equation}
 p(Z_l | V_l, \Mu) \approx \frac{1}{C} p(V_l|Z_l, \Mu) \sum_{z_{N_l(1)}, .., z_{N_l(|N_l|)}}\ \ \  \prod_{m \in \{1, ..., N_l\}} p(Z_l | Z_{N_l(m)}^{(u-1)} = z_{N_l(m)}) p(Z_{N_l(m)}^{(u-1)} = z_{N_l(m)}|V_l, M)
\end{equation}

Factorising the summation over $z_{N_l}$'s we get:

\begin{equation}
 p(Z_l | V_l, \Mu) = p(V_l| Z_l, \Mu)   \prod_{l_2 \in N_l} \sum_{z_{l_2}} p(Z_l | Z_{l_2}^{(u-1)} = z_{l_2}) p(Z_{l_2}^{(u-1)} = z_{l_2}|V_l, M)
\end{equation}

Replacing $z_{l_2}$ with $k_2$ we get:

\begin{equation}
 p(Z_l | V_l, \Mu) = p(V_l| Z_l, \Mu)   \prod_{l_2 \in N_l} \sum_{k_2} p(Z_l | Z_{l_2}^{(u-1)} = k_2) p(Z_{l_2}^{(u-1)} = k_2|V_l, M)
\end{equation}

We shall also denote $z_{lk} = p(Z_l | V_l, \Mu)$. Further simplifications result in:

\begin{equation}
 z_{lk}^{(u)} \propto  \left[ \prod_{i,j \in I} N(V_l^{ij} | f(\alpha_i t_{ij} + \beta_i ; \theta_k), \sigma_k) \right] \left[ \prod_{l_2 \in N_l} \sum_{k_2 = 1}^K z_{l_2k_2}^{(u-1)}\ \Psi (Z_{l} = k, Z_{l_2} = k_2) \right]
\end{equation}

\small
\begin{equation}
 log\ z_{lk}^{(u)} \propto  \left[ \sum_{i,j \in I} -\frac{log\ (2 \pi \sigma_k^2)}{2} - \frac{1}{2\sigma_k^2}(V_l^{ij} - f(\alpha_i t_{ij} + \beta_i ; \theta_k))^2 \right] + \left[\sum_{l_2 \in N_l} log \sum_{k_2 = 1}^K z_{l_2k_2}^{(u-1)} (\delta_{k_2 k}\ exp(\lambda) + (1 - \delta_{k_2 k})\ exp(-\lambda^2)) \right]
\end{equation}
\normalsize

We further define the data-fit term $D_{lk}$ as follows:
\begin{equation}
\label{eq:e-step_Dlk}
D_{lk} = -\frac{log\ (2 \pi \sigma_k^2) |I|}{2} - \sum_{i,j \in I}  \frac{1}{2\sigma_k^2}(V_l^{ij} - f(\alpha_i t_{ij} + \beta_i ; \theta_k))^2 
\end{equation}

This results in:

\small
\begin{equation}
 log\ z_{lk}^{(u)} \propto  D_{lk} + \left[\sum_{l_2 \in N_l} log \sum_{k_2 = 1}^K z_{l_2k_2}^{(u-1)} (\delta_{k_2 k}\ (exp(\lambda) - exp(-\lambda^2)) + exp(-\lambda^2)) \right]
\end{equation}
\normalsize

Finally, we simplify the sum over $k_2$ to get the update equation for $z_{lk}$:

\begin{equation}
\label{eq:e-step}
\begin{split}
 log\ z_{lk}^{(u)} \propto D_{lk}+ \left[ \sum_{l_2 \in N_l} log\ \left[ exp(-\lambda^2) + z_{l_2k}^{(u-1)} (exp(\lambda) - exp(-\lambda^2)) \right] \right]
\end{split}
\end{equation}

In practice, we cannot naively compute the exponential term $z_{lk} = exp(log(z_{lk}))$ due to precision loss. However, we go around this by recomputing the exponentiation and normalisation of $z_{lk}$ simultaneously. Denoting $x(k) = log\ z_{lk}$, for  $k \in [1 \dots K]$, we get: 

\begin{equation}
 z_{lk}  = \frac{e^{x(k)}}{e^{x(1)}+e^{x(2)} + \dots + e^{x(K)}} = \frac{1}{e^{x(1)-x(k)} +e^{x(2)-x(k)} + \dots + e^{x(K)-x(k)}}
\end{equation}

\subsection{M-step}

The M-step itself does not have a closed-form analytical solution. We choose to solve it by successive refinements of the cluster trajectory parameters and the subject time shifts. 

\subsubsection{Optimising trajectory parameters}

\textbf{Trajectory shape - $\theta$}\\

Taking equation \ref{eq:EM1} and fixing the subject time-shifts $\alpha$, $\beta$ and measurement noise $\sigma$, we can find its maximum with respect to $\theta$ only. More precisely, we want:

\begin{equation}
 \theta = \argmax_{\theta} \sum_{z1,\dots, z_L}^K p(Z = (z_1, ..., z_L) | V, \Mu^{(u-1)}) \left[ log\ p(V,Z|M) \right] + log\ p(\theta)
\end{equation}

We observe that for each cluster the individual $\theta_k$'s are conditionally independent, i.e. $\theta_k \ci \theta_m | \{Z, \alpha, \beta, \sigma$\} $\forall k,m$. We also assume that the prior factorizes for each $\theta_k$: $\log p(\theta) = \prod_k^K \log p(\theta_k)$. This allows us to optimise each $\theta_k$ independently:

\begin{equation}
 \theta_k = \argmax_{\theta_k} \sum_{z1,\dots, z_L}^K p(Z = (z_1, ..., z_L) | V, \Mu^{(u-1)}) \left[ log\ p(V,Z|M) \right] + log\ p(\theta_k)
\end{equation}

Replacing the full data log-likelihood, we get:

\begin{equation}
 \theta_k = \argmax_{\theta_k} \sum_{z1,\dots, z_L}^K p(Z = (z_1, ..., z_L) | V, \Mu^{(u-1)})\ log \left[ \prod_{l=1}^L \prod_{(i,j) \in I} N(V_l^{ij} | f(\alpha_i t_{ij} + \beta_i ; \theta_{z_l}), \sigma_{z_l}) \right] + log\ p(\theta_k)
\end{equation}

Note that we didn't include the MRF clique terms, since they are not a function of $\theta_k$. We propagate the logarithm inside the products:

\begin{equation}
 \theta_k = \argmax_{\theta_k} \sum_{z1,\dots, z_L}^K p(Z = (z_1, ..., z_L) | V, \Mu^{(u-1)}) \sum_{l=1}^L \sum_{(i,j) \in I} log\ N(V_l^{ij} | f(\alpha_i t_{ij} + \beta_i ; \theta_{z_l}), \sigma_{z_l}) + log\ p(\theta_k)
\end{equation}

\begin{sloppypar}
We next assume that $Z_l$, the hidden cluster assignment for vertex $l$, is conditionally independent of the other vertex assignments $Z_m$, $\forall m \neq l$ (See E-step approximation from Eq. \ref{eq:e_approx}). This independence assumption induces the following factorization: ${p(Z = (z_1, ..., z_L) | V, \Mu^{(u-1)}) = \prod_l^L p(Z_l = z_l | V, \Mu^{(u-1)})}$. Propagating this product inside the sum over the vertices, we get:
\end{sloppypar}

\begin{equation}
\label{eq:SupTheta5}
 \theta_k = \argmax_{\theta_k} \sum_{l=1}^L \sum_{z_l = 1}^K p(Z_l = z_l | V, \Mu^{(u-1)}) \sum_{(i,j) \in I} log\ N(V_l^{ij} | f(\alpha_i t_{ij} + \beta_i ; \theta_{z_l}), \sigma_{z_l}) + log\ p(\theta_k)
\end{equation}

The terms which don't contain $\theta_k$ dissapear:

\begin{equation}
 \theta_k = \argmax_{\theta_k} \sum_{l=1}^L p(Z_l = k | V, \Mu^{(u-1)}) \sum_{(i,j) \in I} log\ N(V_l^{ij} | f(\alpha_i t_{ij} + \beta_i ; \theta_k), \sigma_k) + log\ p(\theta_k)
\end{equation}

We further expand the gaussian noise model:

\begin{equation}
 \theta_k = \argmax_{\theta_k} \sum_{l=1}^L p(Z_l = k | V, \Mu^{(u-1)}) \sum_{(i,j) \in I} \left[ log\ (2 \pi \sigma_k)^{-1/2} - \frac{1}{2\sigma_k^2}(V_l^{ij} - f(\alpha_i t_{ij} + \beta_i ; \theta_k))^2 \right] + log\ p(\theta_k)
\end{equation}

Constants dissapear due to the $\argmax$ and we get the final update equation for $\theta_k$:

\begin{equation}
 \theta_k = \argmax_{\theta_k} \sum_{l=1}^L p(Z_l = k | V, \Mu^{(u-1)}) \sum_{(i,j) \in I} \left[ - \frac{1}{2\sigma_k^2}(V_l^{ij} - f(\alpha_i t_{ij} + \beta_i ; \theta_k))^2 \right] + log\ p(\theta_k)
\end{equation}

\textbf{Measurement noise - $\sigma$}\\

We first assume a uniform prior on the $\sigma$ parameters to simplify derivations. Using a similar approach as with $\theta$, after propagating the product inside the logarithm and removing the terms which don't contain $\sigma_k$, we get:

\begin{equation}
 \sigma_k = \argmax_{\sigma_k} \sum_{l=1}^L p(Z_l = k | V, \Mu^{(u-1)}) \sum_{(i,j) \in I} log\ N(V_l^{ij} | f(\alpha_i t_{ij} + \beta_i ; \theta_k), \sigma_k)
\end{equation}

Note that, just as for $\theta$ above, the MRF clique terms were not included because they are not a function of $\sigma_k$. Expanding the noise model we get:

\begin{equation}
 \sigma_k = \argmax_{\sigma_k} \sum_{l=1}^L p(Z_l = k | V, \Mu^{(u-1)}) \sum_{(i,j) \in I} \left[ log\ (2 \pi \sigma_k^2)^{-1/2} - \frac{1}{2\sigma_k^2}(V_l^{ij} - f(\alpha_i t_{ij} + \beta_i ; \theta_k))^2 \right]
\end{equation}

The maximum of a function $l(\sigma_k)$ can be computed by taking the derivative of the function $l$ and setting it to zero. This is under the assumption that $l$ is differentiable, which it is but we won't prove it here. This gives:

\begin{equation}
 \frac{\delta l(\sigma_k|.)}{\delta \sigma_k} =  \sum_{l=1}^L p(Z_l = k | V, \Mu^{(u-1)}) \sum_{(i,j) \in I} \frac{\delta}{\delta \sigma_k} \left[ log\ (2 \pi \sigma_k^2)^{-1/2} - \frac{1}{2\sigma_k^2}(V_l^{ij} - f(\alpha_i t_{ij} + \beta_i ; \theta_k))^2 \right]
\end{equation}

Propagating the differential operator further inside the sums we get:

\begin{equation}
 \frac{\delta l(\sigma_k|.)}{\delta \sigma_k} =  \sum_{l=1}^L p(Z_l = k | V, \Mu^{(u-1)}) \sum_{(i,j) \in I} \left[ -\frac{\delta}{\delta \sigma_k} \frac{log\ \sigma_k^2}{2} - \frac{\delta}{\delta \sigma_k} \frac{1}{2\sigma_k^2}(V_l^{ij} - f(\alpha_i t_{ij} + \beta_i ; \theta_k))^2 \right]
\end{equation}

We next perform several small manipulations to reach a more suitable form of the derivative and then set it to be equal to zero:

\begin{equation}
 \frac{\delta l(\sigma_k|.)}{\delta \sigma_k} =  \sum_{l=1}^L p(Z_l = k | V, \Mu^{(u-1)}) \sum_{(i,j) \in I} \left[ - \frac{1}{\sigma_k} - \frac{-2}{2\sigma_k^3}(V_l^{ij} - f(\alpha_i t_{ij} + \beta_i ; \theta_k))^2 \right]
\end{equation}

\begin{equation}
 \frac{\delta l(\sigma_k|.)}{\delta \sigma_k} =  \sum_{l=1}^L p(Z_l = k | V, \Mu^{(u-1)}) \sum_{(i,j) \in I} \left[ - \frac{\sigma_k^2}{\sigma_k^3} + \frac{1}{\sigma_k^3}(V_l^{ij} - f(\alpha_i t_{ij} + \beta_i ; \theta_k))^2 \right]
\end{equation}

\begin{equation}
 \frac{\delta l(\sigma_k|.)}{\delta \sigma_k} =  \sum_{l=1}^L p(Z_l = k | V, \Mu^{(u-1)}) \sum_{(i,j) \in I} \left[ - \sigma_k^2 + (V_l^{ij} - f(\alpha_i t_{ij} + \beta_i ; \theta_k))^2 \right] = 0
\end{equation}


Finally, we solve for $\sigma_k$ and get its update equation: 

\begin{equation}
\label{eq:SupThetaFinal}
 \sigma_k^2 = \frac{1}{|I|} \sum_{l=1}^L p(Z_l = k | V, \Mu^{(u-1)}) \sum_{(i,j) \in I} (V_l^{ij} - f(\alpha_i t_{ij} + \beta_i ; \theta_k))^2
\end{equation}

%

\subsubsection{Estimating subject time shifts - $\alpha$, $\beta$}

For estimating $\alpha$, $\beta$, we adopt a similar strategy as in the case of $\theta$, up to Eq. \ref{eq:SupTheta5}. This gives us the following problem:

\begin{equation}
 \alpha_i, \beta_i = \argmax_{\alpha_i, \beta_i} \sum_{l=1}^L \sum_{k = 1}^K p(Z_l = k | V, \Mu^{(u-1)}) \sum_{(i',j) \in I} log\ N(V_l^{i'j} | f(\alpha_{i'} t_{i'j} + \beta_{i'} ; \theta_{k}), \sigma_{k}) + log\ p(\alpha_i, \beta_i)
\end{equation}

The terms $\alpha_{i'}, \beta_{i'}$ for other subjects $i' \neq i$ dissappear:

\begin{equation}
 \alpha_i, \beta_i = \argmax_{\alpha_i, \beta_i} \sum_{l=1}^L \sum_{k = 1}^K p(Z_l = k | V, \Mu^{(u-1)}) \sum_{j \in I_i} log\ N(V_l^{ij} | f(\alpha_{i} t_{ij} + \beta_{i} ; \theta_{k}), \sigma_{k}) + log\ p(\alpha_i, \beta_i)
\end{equation}

Expanding the gaussian noise model we get:

\begin{equation}
 \alpha_i, \beta_i = \argmax_{\alpha_i, \beta_i} \sum_{l=1}^L \sum_{k = 1}^K p(Z_l = k | V, \Mu^{(u-1)}) \sum_{j \in I_i} \left[ log\ (2 \pi \sigma_k^2)^{-1/2} - \frac{1}{2\sigma_k^2}(V_l^{ij} - f(\alpha_i t_{ij} + \beta_i ; \theta_k))^2 \right]  + log\ p(\alpha_i, \beta_i)
\end{equation}

After removing constant terms we end up with the final update equation for $\alpha_i$, $\beta_i$:

\begin{equation}
 \alpha_i, \beta_i = \argmin_{\alpha_i, \beta_i}  \left[ \sum_{l=1}^L \sum_{k=1}^K p(Z_l = k | V, \Mu^{(u-1)}) \frac{1}{2\sigma_k^2} \sum_{j \in I_i} (V_l^{ij} - f(\alpha_i t_{ij} + \beta_i ; \theta_k))^2\right] - log\ p(\alpha_i, \beta_i)
\end{equation}

\subsubsection{Estimating MRF clique term - $\lambda$}

We optimise $\lambda$ using the following formula:

$$\lambda^{(u)} = \argmax_{\lambda} E_{p(Z|V, M^{(u-1)}, \lambda, Z^{(u-1)})}[log\ p(V,Z|M^{(u-1)})] $$

Note that $p(Z|V, M^{(u-1)}, \lambda, Z^{(u-1)})$ is a function of $\lambda$, so for each lambda we estimate $z_{lk}$ through approximate inference. We do this because otherwise the optimisation of $\lambda$ will only take into account the clique terms and completely exclude the data terms. We further simplify the objective function for lambda as follows:

\small
$$\lambda^{(u)} = \argmax_{\lambda} \sum_{z1,\dots, z_L}^K p(Z = (z_1, ..., z_L) | V, \Mu^{(u-1)}, \lambda, Z^{(u-1)})\ log \left[ \prod_{l=1}^L \prod_{(i,j) \in I} N(V_l^{ij} | f(\alpha_i t_{ij} + \beta_i ; \theta_{z_l}), \sigma_{z_l}) \prod_{l = 1}^{L}\prod_{l_2 \in N_l} \Psi (z_{l}, z_{l_2}) \right] $$
\normalsize

We take the logarithm:

$$\lambda^{(u)} = \argmax_{\lambda} \sum_{z1,\dots, z_L}^K p(Z = (z_1, ..., z_L) | V, \Mu^{(u-1)}, \lambda, Z^{(u-1)})\ \left[ \sum_{l=1}^L \sum_{(i,j) \in I} log\ N(V_l^{ij} | ..) + \sum_{l = 1}^{L}\sum_{l_2 \in N_l} log\ \Psi (z_{l}, z_{l_2}) \right] $$

Let us denote $z_{lk} = p(Z_l = k | V, \Mu^{(u-1)}, \lambda, Z^{(u-1)})$. Assuming independence between the latent variables $Z_l$ we get:

\begin{multline}
\lambda = \argmax_{\lambda} \sum_{l=1}^L \sum_{k=1}^K z_{lk}\ \left[ \sum_{(i,j) \in I} log\ N(V_l^{ij} | ..) \right] + \sum_{l = 1}^{L} \sum_{k=1}^K z_{lk} \sum_{l_2 \in N_l} \sum_{k_2 = 1}^K z_{l_2k}\ log\ \Psi (Z_{l} = k, Z_{l_2} = k_2)  
\end{multline}

However, we now want to make $z_{lk}$ a function of $\lambda$ as previously mentioned, so $z_{lk}=\zeta_{lk}(\lambda)$, for some function $\zeta_{lk}$. More precisely, using the E-step update from Eq. \ref{eq:e-step} we define for each vertex $l$ and cluster $k$ a function $\zeta_{lk}(\lambda)$ as follows:

$$\zeta_{lk}(\lambda) = exp \left( D_{lk} +   \sum_{l_2 \in N_l} log\ \left[ exp(-\lambda^2) + z_{l_2k}^{(u-1)} (exp(\lambda) - exp(-\lambda^2)) \right] \right)$$

where $D_{lk}$ is as defined in Eq \ref{eq:e-step_Dlk}. We then replace $z_{lk}$ with $\zeta_{lk}(\lambda)$ and introduce the chosen MRF clique model to get:



$$ \lambda^{(u)} = \argmax_{\lambda}\ \sum_{l=1}^L \sum_{k=1}^K \zeta_{lk}(\lambda) D_{lk}  + \sum_{l = 1}^{L} \sum_{k}^K \sum_{l_2 \in N_l} \sum_{k_2 = 1}^K \zeta_{lk}(\lambda) \zeta_{l_2k}(\lambda)\ \left[ \delta_{kk_2} \lambda + (1-\delta_{kk_2}) (-\lambda^2)\right]  $$

We separate the cliques that have matching clusters to the ones that don't:

$$ \lambda^{(u)} = \argmax_{\lambda}\ \sum_{l=1}^L \sum_{k=1}^K \zeta_{lk}(\lambda) D_{lk} + \sum_{l = 1}^{L} \sum_{l_2 \in N_l} \sum_{k}^K  \left[ \zeta_{lk}(\lambda) \zeta_{l_2k}(\lambda)\ \lambda + \sum_{k2 \neq k} \zeta_{lk}(\lambda) \zeta_{l_2k}(\lambda) (-\lambda^2)\right]  $$

We also factorise the clique terms:


$$ \lambda^{(u)} = \argmax_{\lambda}\ \sum_{l=1}^L \sum_{k=1}^K \zeta_{lk}(\lambda) D_{lk} + \lambda \sum_{l = 1}^{L} \sum_{l_2 \in N_l} \sum_{k}^K   \zeta_{lk}(\lambda) \zeta_{l_2k}(\lambda)\  + (-\lambda^2) \sum_{l = 1}^{L} \sum_{l_2 \in N_l} \sum_{k}^K \zeta_{lk}(\lambda) (1- \zeta_{l_2k}(\lambda))   $$

Finally, we simplify to get the objective function for $\lambda$.

\begin{equation}
 \lambda^{(u)} = \argmax_{\lambda}\ \sum_{l=1}^L \sum_{k=1}^K \zeta_{lk}(\lambda) \left[  D_{lk} \  + \lambda \sum_{l_2 \in N_l}  \zeta_{l_2 k}(\lambda)\  -\lambda^2 \sum_{l_2 \in N_l} (1- \zeta_{l_2 k}(\lambda))  \right]
\end{equation}

For implementation speed-up, data-fit terms $D_{lk}$ can be pre-computed.

\section{Fast DIVE Implementation - Proof of Equivalence}

Fitting DIVE can be computationally prohibitive, especially given that the number of vertices/voxels can be very high, e.g. more than 160,000 in our datasets. We derived a fast implementtion of DIVE, which is based on the idea that for each subject we compute a weighted mean of the vertices within a particular cluster, and then compare that mean with the corresponding trajectory value. This is in contrast with comparing the value at each vertex with the corresponding trajectory of its cluster. In the next few sections, we will present the mathematical formulation of the fast implementation for parameters [$\theta$, $\alpha$, $\beta$]. Parameter $\sigma$ already has a closed-form update, while parameter $\lambda$ has a more complex update procedure for which this fast implementation doesn't work. For each parameter, we will also provide proofs of equivalence.

\subsection{Trajectory parameters - $\theta$}

\subsubsection{Fast implementation}

The fast implementation for $\theta$ implies that, instead of optimising Eq. \ref{eq:SupThetaFinal} we optimise the following problem:
\begin{equation}
\label{eq:supThetaFast1}
 \theta_k = \argmin_{\theta_k} \sum_{(i,j) \in I} (<V^{ij}>_{\hat{Z}_k} - f(\alpha_i t_{ij} + \beta_i ; \theta_k))^2
\end{equation}

where $<V^{ij}>_{\hat{Z}_k}$ is the mean value of the vertices belonging to cluster $k$. Mathematically, we define $\hat{Z}_k = [z_{1k}\gamma_k,\ z_{2k}\gamma_k,\ \dots,\ z_{Lk}\gamma_k ]$ where  $\gamma_k = (\sum_{l=1}^L z_{lk})^{-1}$ is the normalisation constant. Moreover, we have that ${<V^{ij}>_{\hat{Z}_k} = \sum_{l=1}^L z_{lk} \gamma_k V^{ij}}$. We take the derivative of the likelihood function $l_{fast}$ of the fast implementation (Eq. \ref{eq:supThetaFast1}) with respect to $\theta_k$ and perform several simplifications:
\begin{equation}
\frac{\delta l_{fast}(\theta_k|.)}{\delta \theta_k} = \frac{\delta}{\delta \theta_k} \sum_{(i,j) \in I} \left( \sum_{l=1}^L z_{lk} \gamma_k V^{ij} - f(\alpha_i t_{ij} + \beta_i ; \theta_k) \right)^2
\end{equation}

\begin{equation}
\frac{\delta l_{fast}(\theta_k|.)}{\delta \theta_k} = \sum_{(i,j) \in I} 2 \left( \left( \sum_{l=1}^L \gamma_k z_{lk} V^{ij} \right) - f(\alpha_i t_{ij} + \beta_i ; \theta_k) \right) \frac{-\delta f(.)}{\delta \theta_k}
\end{equation}

using the fact that $\sum_{l=1}^L \gamma_k z_{lk} = 1$ we get:

\begin{equation}
\frac{\delta l_{fast}(\theta_k|.)}{\delta \theta_k} = \sum_{(i,j) \in I} 2 \left( \sum_{l=1}^L \gamma_k z_{lk} \left( V^{ij} - f(\alpha_i t_{ij} + \beta_i ; \theta_k) \right) \right) \frac{-\delta f(.)}{\delta \theta_k}
\end{equation}

\begin{equation}
\frac{\delta l_{fast}(\theta_k|.)}{\delta \theta_k} = 2 \gamma_k \sum_{(i,j) \in I} \frac{-\delta f(.)}{\delta \theta_k} \left( \sum_{l=1}^L z_{lk} \left( V^{ij} - f(\alpha_i t_{ij} + \beta_i ; \theta_k) \right) \right)
\end{equation}

By setting the derivative to zero, the optimal $\theta$ is thus a solution of the following equation:

\begin{equation}
\label{eq:supThetaFast2}
\sum_{(i,j) \in I} \frac{-\delta f(.)}{\delta \theta_k} \left( \sum_{l=1}^L z_{lk} \left( V^{ij} - f(\alpha_i t_{ij} + \beta_i ; \theta_k) \right) \right) = 0
\end{equation}

\subsubsection{Slow implementation}

We will prove that if theta is a solution of the slow implementation, it is also a solution of Eq. \ref{eq:supThetaFast2}, which will prove that the fast implementation is equivalent. The slow implementation is finding $\theta$ from the following equation:

\begin{equation}
 \theta_k = \argmin_{\theta_k} \sum_{l=1}^L z_{lk} \sum_{(i,j) \in I} (V_l^{ij} - f(\alpha_i t_{ij} + \beta_i ; \theta_k))^2
\end{equation}

Taking the derivative of the function above ($l_{slow}$) with respect to $\theta_k$ we get:
\begin{equation}
\frac{\delta l_{slow}(\theta_k|.)}{\delta \theta_k} = \sum_{l=1}^L z_{lk} \sum_{(i,j) \in I} 2 (V_l^{ij} - f(\alpha_i t_{ij} + \beta_i ; \theta_k)) \left(-\frac{\delta f(.)}{\delta \theta_k} \right) = 0
\end{equation}

After swapping terms around and using distributivity we get:

\begin{equation}
  \sum_{(i,j) \in I} \left(-\frac{\delta f(.)}{\delta \theta_k} \right) \sum_{l=1}^L z_{lk} (V_l^{ij} - f(\alpha_i t_{ij} + \beta_i ; \theta_k))  = 0
\end{equation}

This is the same optimisation problem as in Eq. \ref{eq:supThetaFast2}, which proves that the two formulations are equivalent with respect to $\theta$.

\subsection{Noise parameter - $\sigma$}

The noise parameter $\sigma$ can actually be computed in a closed-form solution for the original slow model implementation, so there is no benefit in implementing the fast update for $\sigma$. Moreover, the $\sigma$ in the fast implementation computed the standard deviation in the \emph{mean value} of the vertices within a certain cluster, and not the deviation withing the \emph{actual value} of the vertices.

\subsection{Subjects-specific time shifts - $\alpha$, $\beta$}

\subsubsection{Fast implementation}

The equivalent fast formulation for the subject-specific time shifts is similar to the one for the trajectory parameters. It should be noted however that we need to weight the sums corresponding to each cluster by $\gamma_{k}^{-1}$. This gives the following equation for the fast formulation:

\begin{equation}
 \alpha_i, \beta_i = \argmin_{\alpha_i, \beta_i}  \sum_{k=1}^K \gamma_k^{-1} \frac{1}{2\sigma_k^2} \sum_{j \in I_i} (<V_l^{ij}>_{\hat{Z}_k} - f(\alpha_i t_{ij} + \beta_i ; \theta_k))^2 = 0
\end{equation}

In order to prove that this is equivalent to the slow version, we need to take the derivative of the likelihood function ($l_{fast}$) from the above equation with respect to $\alpha_i$, $\beta_i$ and set it to zero:

\begin{equation}
 \frac{\delta l_{fast}(\alpha_i, \beta_i|.)}{\delta \alpha_i, \beta_i} =  \frac{\delta}{\delta \alpha_i, \beta_i} \sum_{k=1}^K \gamma_k^{-1} \frac{1}{2\sigma_k^2} \sum_{j \in I_i} (<V_l^{ij}>_{\hat{Z}_k} - f(\alpha_i t_{ij} + \beta_i ; \theta_k))^2 = 0
\end{equation}

We expand the average across the vertices and slide the derivative operator inside the sums:

\begin{equation}
 \sum_{k=1}^K \gamma_k^{-1} \frac{1}{2\sigma_k^2} \sum_{j \in I_i} 2 \left( \sum_{l=1}^L \gamma_k z_{lk} V_l^{ij} - f(\alpha_i t_{ij} + \beta_i ; \theta_k) \right) \frac{-\delta f(.)}{\delta \alpha_i, \beta_i}
\end{equation}

Since $ \sum_{l=1}^L \gamma_k z_{lk} = 1$ we get:

\begin{equation}
 2 \sum_{k=1}^K \gamma_k^{-1} \frac{1}{2\sigma_k^2} \sum_{j \in I_i} \frac{-\delta f(.)}{\delta \alpha_i, \beta_i}  \left( \sum_{l=1}^L \gamma_k z_{lk} (V_l^{ij} - f(\alpha_i t_{ij} + \beta_i ; \theta_k)) \right) 
\end{equation}

Removing the factor 2 and sliding $\gamma_k$:

\begin{equation}
 \sum_{k=1}^K \gamma_k^{-1} \gamma_k \frac{1}{2\sigma_k^2} \sum_{j \in I_i} \frac{-\delta f(.)}{\delta \alpha_i, \beta_i}  \left( \sum_{l=1}^L  z_{lk} (V_l^{ij} - f(\alpha_i t_{ij} + \beta_i ; \theta_k)) \right) 
\end{equation}

Further sliding $\sum_{l=1}^L z_{lk}$ to the left we get the final optimisation problem:
\begin{equation}
\label{eq:supAlphaFast2}
 \sum_{k=1}^K  \frac{1}{2\sigma_k^2} \sum_{l=1}^L z_{lk} \sum_{j \in I_i} \frac{- \delta f(.)}{\delta \alpha_i, \beta_i} (V_l^{ij} - f(\alpha_i t_{ij} + \beta_i ; \theta_k))
\end{equation}

\subsubsection{Slow implementation}

In a similar way to the trajectory parameters, we want to prove that solving the problem from Eq. \ref{eq:supAlphaFast2} (fast implementation) is the same as solving the original slow implementation problem, which is defined as:

\begin{equation}
 \alpha_i, \beta_i = \argmin_{\alpha_i, \beta_i}   \sum_{l=1}^L \sum_{k=1}^K z_{lk} \frac{1}{2\sigma_k^2} \sum_{j \in I_i} (V_l^{ij} - f(\alpha_i t_{ij} + \beta_i ; \theta_k))^2
\end{equation}

Taking the derivative of the function above with respect to $\alpha_i, \beta_i$, we get:

\begin{equation}
 \frac{\delta l_{slow}(\alpha_i, \beta_i|.)}{\delta \alpha_i, \beta_i} = \sum_{k=1}^K \frac{1}{2\sigma_k^2} \sum_{l=1}^L z_{lk} \sum_{j \in I_i} \frac{- \delta f(.)}{\delta \alpha_i, \beta_i} (V_l^{ij} - f(\alpha_i t_{ij} + \beta_i ; \theta_k))
\end{equation}

This is the same problem as the fast implementation one from Eq. \ref{eq:supAlphaFast2}, thus the fast model is equivalent to the slow model with respect to $\alpha$, $\beta$.

\end{document}